\documentclass[mathfont=12]{uai2023}

\usepackage[american]{babel}

\usepackage{natbib} 
\usepackage{mathtools} 
\usepackage{booktabs} 
\usepackage{tikz} 
\usepackage{lipsum,graphicx,multicol}
\usepackage{xcolor}
\usepackage{subfigure}
\usepackage{mathtools}
\usepackage{amsthm}
\usepackage{url}
\usepackage{hyperref}
\usepackage{verbatim}
\usepackage{multirow}

\title{Stress and Adaptation: Applying Anna Karenina Principle in Deep Learning for Image Classification}

\author[1]{\href{mailto:<nesma.mahmoud@ut.ee>?Subject=Your UAI 2023 paper}{Nesma Mahmoud}{}}
\author[2]{Hanna Antson}
\author[3]{jaesik choi}
\author[2]{Osamu Shimmi}
\author[1]{Kallol Roy}
\affil[1]{
    Institute of Computer Science\\
    University of Tartu\\
    Estonia
}
\affil[2]{
    Institute of Molecular and Cell Biology\\
    University of Tartu\\
    Estonia
}
\affil[3]{
    School of Artificial Intelligence\\
    Korea Advanced Institute of Science and Technology\\
    Korea
  }
  
\begin{document}
\maketitle
\begin{abstract}
Image classification with deep neural networks  has reached state-of-art with high accuracy. This success is attributed to good internal representation (features) that bypasses the difficulties of the non-convex optimization problems. We have little understanding of these internal representations, let alone quantifying them. Recent research efforts have focused on alternative theories and explanations of the generalizability of these deep networks. We propose the alternative: perturbation of deep models during their training induces changes that lead to transitions to different families. The result is an ‘Anna Karenina Principle (AKP)’ for deep learning, in which less generalizable models (unhappy families) vary more in their representation than more generalizable models (happy families)—paralleling Leo Tolstoy’s dictum that “all happy families look alike, each unhappy family is unhappy in its own way.” Anna Karenina principle has been found in systems in a wide range: from the surface of endangered corals exposed to harsh weather to the lungs of patients suffering from fatal diseases of AIDs. In our paper, we have generated artificial perturbations to our model by hot-swapping the activation and loss functions during the training. In this paper, we build a model to classify cancer cells from non-cancer ones. We give theoretical proof that the internal representations of generalizable (happy) models are similar in the  asymptotic limit. Our experiments verify similar representations of generalizable models.
 
\end{abstract}

\section{Introduction}\label{sec:intro}
 The spectacular success of deep learning in vision (language) is attributed to the essential features learned by them. These deep neural networks build intermediate representations~\cite{rumelhart:errorpropnonote}. These deep neural networks are trained end-to-end to build their intermediate representations for a richer representation~\cite{goh2021multimodal, olah2017feature}. However, the formal characterization of these representations is extremely hard. Different training methods for good training and evaluation accuracy are used to support the  existence of good representations of the data inside the  model. But we believe that not only the training but architecture, objective function, and other factors contribute to developing good representations. So, highly accurate models have better intermediate representations than their low-accurate counterparts. Thus the better representation comes from the evolutionary principle~\cite{west2003developmental,diamond98}: a deep net model adapts by correspondingly building its optimal representation to survive (a.k.a  have good accuracy). We observe through our experiments Anna Karenina Principle (AKP) all “good representations” are essentially similar and all “bad representations” are bad in their own way~\cite{DBLP:journals/corr/abs-2106-07682,osti_10218957, https://doi.org/10.48550/arxiv.2209.04836}. Recently some research groups have connected the notion of symmetry with models' internal representation ~\cite{https://doi.org/10.48550/arxiv.2205.14258}. We do not have sufficient understanding and good metrics to measure the similarity between good representations. The high test accuracy  is an efficient heuristic  for building good intermediate representations for the Anna Karenina scenario. Deep learning-based vision model tasks will have internal representations that focus on the curves 
and sharp edges, while large language models focus on the distributional properties of the text. Thus the ubiquitous loss surface  with exponentially many local optima separated by numerous barriers is not a good metric to infer the intermediate representations
~\cite{NEURIPS2018_a41b3bb3, szegedy2015rethinking}.  For AKP the following emerges~\cite{DBLP:journals/corr/abs-2106-07682,Gorban_2010,doi:10.1177/0971333616689398}:
\begin{itemize}
  \item All successful models (high test accuracy) have similar internal representations.
  \item Better internal representations are built with more data,  bigger model sizes, and more computing.
  \item  Internal representations are built from evolutionary principles; irrelevant features are dropped on the way for the  models to survive.
\end{itemize}
The organization of the paper is as follows: In Section 2, background notions and related work are introduced. The theory is proven in Section 3, and the design of the experiments is presented in Section 4. Section 5 displays the dataset setup, and Section 6 offers an empirical evaluation. The paper concludes in Section VI.
\section{Anna Karenina Principle (AKP)}
According to the Wikipedia definition, Anna Karenina principle (AKP) states that ''a deficiency in any one of several factors dooms an endeavor to failure''~\cite{wiki:Anna}. This principle reveals the intrinsic mechanism of adaptation to crises, from human adaptation to harsh conditions to the dotcom bubble burst to shift animal microbiomes from a healthy to a dysbiotic stable state~\cite{diamond98,osti_10218957}. The principle derives its name from Leo Tolstoy's famous novel Anna Karenina: ''All happy families are alike; each unhappy family is unhappy in its own way''.  This translates in our domain to successful (a.k.a) generalizable) models share a common set of attributes (a.k.a features) that lead to success (high test accuracy).
In contrast, a deficiency of various attributes (a.k.a features) can cause an unhappy (low test accuracy) family. This concept is valid in several research fields. With the mathematical framework of correlation and variance of multiple systems (agents) facing external stress, we can predict the symptoms of crisis by observing the increase of correlation among the agents~\cite{doi:10.1177/0971333616689398}.
 \begin{figure}[htp]
    \centering
\includegraphics[width=0.50\textwidth]{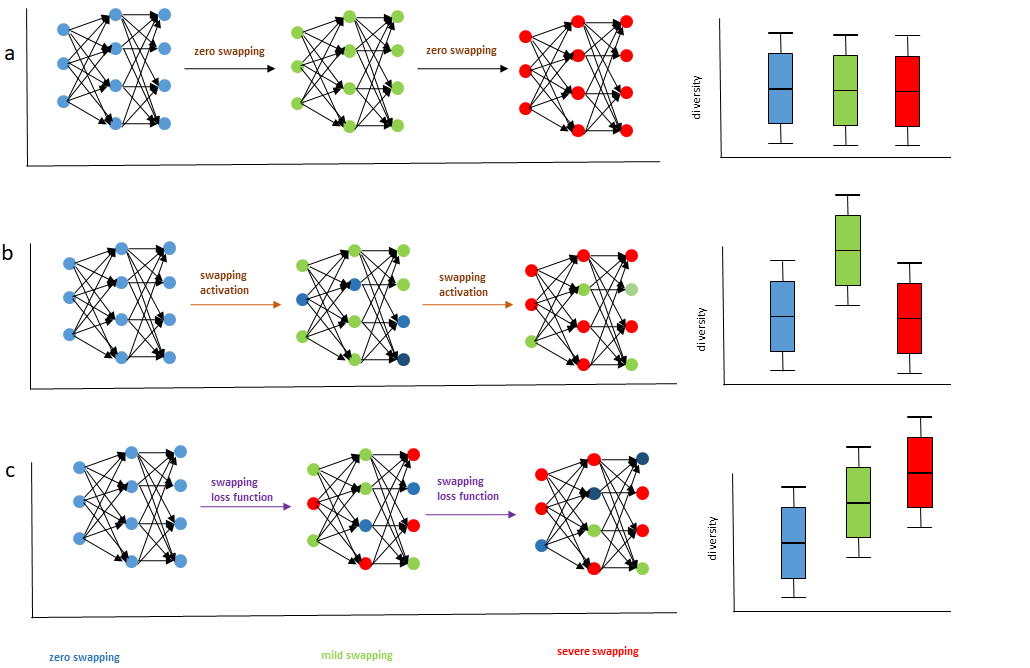}
    \caption{Anna Karenina Principle}
    \label{Anna Karenina Principle}
\end{figure}
We show a hypothetical ordination of alteration of neuron sample density because of the  perturbation from swapping the activation and loss functions during the training shown in Figure \ref{Anna Karenina Principle}~\cite{osti_10218957}. There are three colors to label the  neurons (i) blues spheres samples represent neurons under no stress (perturbations) or baseline neurons (ii)  green spheres samples represent neurons under mild stress (iii) red spheres samples represent neurons under severe stress.  The X-axis represents the increasing  stress (perturbations) given to the model by swapping methods. These perturbations produce clusters of ‘stressed’ neuron samples (green and red spheres) from the initial (a.k.a unstressed) neuron samples (blue spheres). Figure \ref{Anna Karenina Principle}.a) shows under a linear increase in stress, the neuronal sample changes deterministically, from blue (no stress) to  green (mild stress) to red (severe stress). Figure \ref{Anna Karenina Principle}.b) shows when the stress (perturbation) is non-linear through the change of swapping activation functions. This produces clusters of healthy (blue
spheres) and severely stressed (red spheres) samples. This results in a change of neuron samples from no-stress neurons (blue spheres) to a mix of green and blue clusters (in mild stress)  to a mix of red and green spheres (in severe stress). Similar is Figure \ref{Anna Karenina Principle}.c) that shows non-linear perturbations through swapping of loss functions changes neuron sample density. The correlation coefficient  between stress neuron samples (green and red spheres) and zero-stress neuron samples (blue spheres) is computed.

\section{AKP in Image Classification}
Image classification with deep neural networks took off from ImageNet moment~\cite{krizhevsky2017imagenet}. Series of spectacular discoveries of 
Capsule-net, ResNet50, MobileNet, and recently Image Transformer mark the golden moment of computer vision~\cite{NIPS2017_2cad8fa4,he2016residual, howard2017mobilenets}. A significant part of computer vision research goes into medical research, particularly in image segmentation, classification, imaging etc. Das et al. used transfer learning from the ResNet50 feature extraction model to detect Acute Lymphoblastic Leukemia (ALL) from a small medical dataset~\cite{das2021detection}. Choudhary et al. used transfer learning for breast cancer classification with  three pre-trained CNNs models, ResNet34, VGG19, and ResNet50~\cite{choudhary2021transfer}. In a recent study of breast cancer intensity spread by cancer, Ghrabat et al. used a convolution neural network(CNN) from scratch and with data augmentation. ~\cite{ghrabat2022fully}. The deep neural models perform well on the generalizability to out of distributions because of optimal feature selection from the image data. Deep neural models capture relevant features (attributes) from the image data, e.g., ridge crease, valley crease, boundary edge, marking curve, corner, convex corner, concave corner, saddle corner, notch corner~\cite{10.5555/3017671, Hartley:2003:MVG:861369}.

\subsection{Proof}
We employ a model (inception network) for binary classification (cancer or no cancer). The model is trained on the dataset $D$, with the input $x$ as the image (from microscopy), with training labels $y \in \{ 0, 1\}$ distributed as $p(y|x, D)$. It is trained with the stochastic gradient descent (SGD) and RMSprop using binary cross entropy loss function. The predicted labelled output is~\cite{murphy2018machine}:
\begin{equation}
\hat{y} =  \arg \max_{c = 1}^{2} p( y = c|x, D,  W_{f})
\end{equation}
where $W_{f}$ is the final trained weight after epoch $E$. The initial weight tensor is $W_{0}$ (sampled from random and glorot initialization). We use perturbation training dynamics, where we discretize the epoch in discrete intervals $E = \{ E_{0} \cup  E_{2} \cdots E_{2}  \}$ where at different interval $E_{i}$ we use a different activation function (e.g., Relu, Tanh) or loss functions. Online changing the activation functions is a perturbation given to the model as it resets (partially) the training. AKP Principle hypothesizes models (biological and non-biological agents) going through similar training dynamics (and environmental) have similar internal representations. This inspires researchers to experiment with \textit{model stitching} where  some layers of a model are stitched inside another model ~\cite{https://doi.org/10.48550/arxiv.1411.5908, DBLP:journals/corr/abs-2106-07682}. This shows the reason why solving  non-convex neural optimization network loss landscapes succeeds: the presence of a single basin after considering all possible permutation symmetries of hidden layers ~\cite{https://doi.org/10.48550/arxiv.2209.04836}. 

In our paper, we propose a new theory for the Anna Karenina Principle (AKP) emergence. We draw the tools from evolutionary game theory and population dynamics to model~\cite{books/cu/HofbauerS98, RePEc:eee:jetheo:v:77:y:1997:i:1:p:1-14}. We model the artificial neuron population as a population of agents (similar to biological species, e,g animals and insects). The population frequencies of different groups of neurons in our model web complex interactions, from which the 
 feature vectors emerge. The complex interactions between different groups of neurons (inter-species competition) or among the neurons within the same group (intra-species competition). Different groups of neurons were fighting for their survival and are modeled as the celebrated  predator-prey equations of Lotka–Volterra equations. In addition to back-propagation, this predator-prey interaction dynamics updates the neuron weights.  The online swapping of activations and loss functions enhances the evolutionary pressure among the neurons. The features from the data extracted come from the evolutionary pressure of the survivability of the neurons. For simplicity, we model our neuron population of size two $W_{1}$ and $W_{2}$, where $W_{1}$ is the prey and $W_{2}$ is the predator, evolved through Lotka–Volterra equations:
 
\begin{equation}\label{Lotka–Volterra}
\begin{aligned}
\dot{W_{1}} &=  W_{1}(a - bW_{2})  \\
\dot{W_{2}} &=  W_{2}(-c + dW_{1}) 
\end{aligned}
\end{equation}
where $\dot{W_{1}} = \frac{W_{1}}{dt}$, $\dot{W_{2}} = \frac{W_{2}}{dt}$ and constants  $a, b, c, d \geq 0$. Populations $W_{1}$ and $W_{1}$ learn the features (e.g., corners, bends, etc.) of the image. For example, the population $W_{1}$ learns sharp corners, and population $W_{2}$ learns more complex features. The mapping between neuron population and its learning features capacity is a complex problem and poorly understood. Though we propose such a map, $T$ exists, and it is differentiable. Again we take two image feature set $F = \{f_{1}, f_{2} \}$ for simplicity. These features are dependent on the population of neurons as a linear combination of populations as:

 \begin{gather}\label{feature}
 \begin{bmatrix} f_{1}  \\ f_{2}  \end{bmatrix}
 =
  \begin{bmatrix}
   T_{11}    &
   X_{12}    \\
   T_{21}    &
   T_{22}    
   \end{bmatrix}
\begin{bmatrix} W_{1}  \\ W_{2}  \end{bmatrix}
\end{gather}

Solving time-independent Lotka–Volterra Equation\ref{Lotka–Volterra} we and plugging in Equation~\ref{feature} we get:

 \begin{gather}\label{feature}
 \begin{bmatrix} f_{1}(t)  \\ f_{2}(t)  \end{bmatrix}
 =
  \begin{bmatrix}
   T_{11}    &
   X_{12}    \\
   T_{21}    &
   T_{22}    
   \end{bmatrix}
\begin{bmatrix} W_{1}(0)e^{at}  \\ W_{2}(0)e^{-ct}  \end{bmatrix}
\end{gather}

From Equation~\ref{feature}, we infer the feature vectors are functions of time that depend on the population of neurons $W_{1}, W_{2}$. So survivability of neuron populations is projected to the feature survivability. This we term as \textit{feature fight}, and thus features are not extracted only through gradient descent but also from predator-prey equations of Lotka–Volterra equations. The features extracted by different models in which their respective neurons have gone through similar fights for survival are similar, echoing the Anna Karenina Principles.

\section{Design of Experiments}
We design the model's training from the evolutionary principle of \textit{genocopy phenocopy interchangeability} ~\cite{west2003developmental}. We study the learning dynamics of the models with stochastic perturbations ~\cite{KANIOVSKI1995330}. The perturbations are induced through the swapping of activation and loss functions during training. This swapping of hyperparameters resets the training process and thus puts pressure evolutionary on the model to survive. In our experiment design, we have used Inceptionv3, a convolutional neural network (CNN), for binary classification. The features are extracted hierarchically, with low-level features first and more complex features later with optimization. We use pre-trained ImageNet weights for our binary classification task and thus bypassing the requirements for large sample complexity for training. We run our experiments separately with two different optimizers (i) Stochastic Gradient Descent (SGD) and (ii) Root Mean Square Propagation (RMS-Prop). We report three sets of experiments in our paper: 
\begin{enumerate}
  \item \textbf{Group A:} Training Inceptionv3 models for 30 epochs with a Fixed seed.
  \item \textbf{Group B:} Training Inceptionv3 models for 30 epochs with a Random seed.
  \item  \textbf{Group C:} Training Inceptionv3 models for 100 epochs with different initializers.
\end{enumerate}

\subsection{Hot Swapping Activation Functions}
 Hot swapping activation functions is a dynamic technique in which the activation functions of the model are changed during training and monitoring the network's performance to search for the best activation function for each layer. In our model, we incorporated three dense layers, with the final layer having a softmax activation function for classifying the model's classes. While training our model, we experimented with switching the activation functions of the first two dense layers. We choose different activation functions - Tanh, Softplus, and ReLU. Our policy of changing (swapping)  activation functions at epoch numbers:  3, 6, 9, 12, 15, 18, and 21.  The softmax activation function at the last layer is not changed as the final classification task depends on it.

\subsection{Interchangeability of Loss Functions}
We swap (interchange) three different loss  functions - Poisson, Kullback-Leibler divergence (KLDivergence), and Sparse categorical cross-entropy during the training for a period of 30 epochs. Our policy of interchanging loss functions happens at epochs 3, 6, 9, 12, 15, 18, and 21. We kept track of both the training loss and validation loss over the course of 30 epochs to observe how changing the loss function impacted the model's performance. Those loss functions were interchangeably measuring the discrepancy between the model's predictions and the true labels to help guide the optimization process.

\noindent \section{Dataset Setup} 
\begin{figure}
\centering
 \subfigure[]{\includegraphics[width=0.3\linewidth]{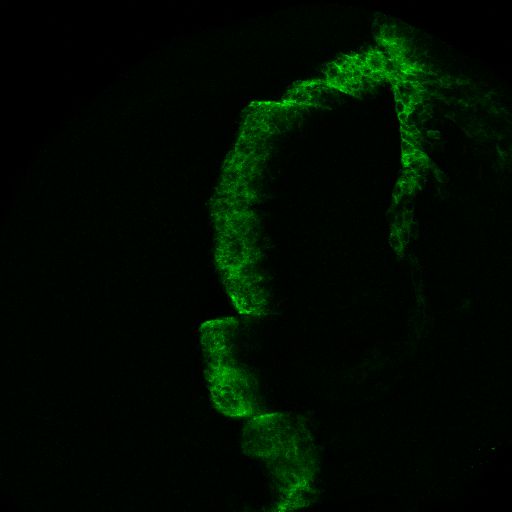}}
 \subfigure[]{\includegraphics[width=0.3\linewidth]{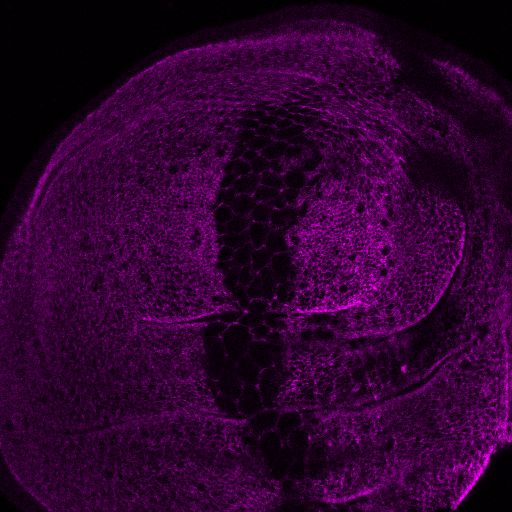}}
 \subfigure[]{\includegraphics[width=0.3\linewidth]{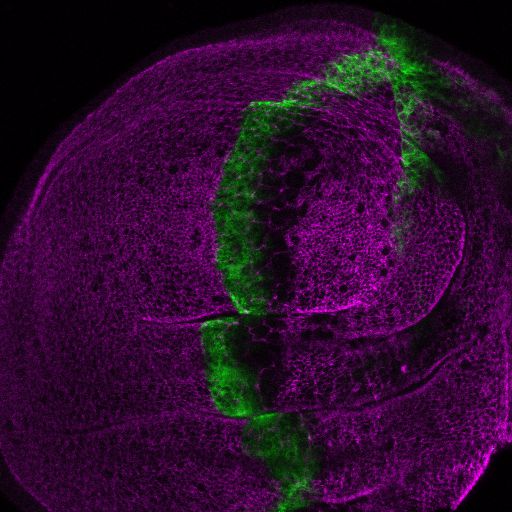}}
 \caption{(a) Stacked Green  channel (b) Stacked Magenta channel (c) The merged channels}
    \label{fig:channels}
\end{figure}

The dataset \ref{Dataset} in this study consists of experimental studies carried out in the lab. The data is of two classes: images of flies that will develop tumors (cancer) and images of flies that will not develop tumors (cancer). The data set is imbalanced, and fewer images are  with tumors than those without tumors. Our study's results will additionally help to improve our understanding of tumor development in flies and may have implications for cancer research in humans. Microscopy images are special kinds of 3D images that capture the internal structure of a specimen. Each image has three dimensions: width, height, and depth as shown in Figure~\ref{fig:channels}. The width and height are represented by the x and y-axes, while the z-axis represents the depth. The depth information is particularly useful for capturing multiple layers of the specimen, such as the tissue layers of a wing, which can have up to 13 layers. Additionally, each layer is comprised of three different color channels. The layers and color channels are combined into a single, stacked image to make it easier for a neural network to process these images. This is done by taking the z-axis values for each color channel and merging them into two separate 2D images, one for the green channel and one for the magenta channel. These two images are then combined to form the final 3D image, which contains all the depth and color information of the wing tissue. So, to put it simply, microscopy images are complex 3D images of a specimen's internal structure, and they are processed into a single stacked image to simplify their use as input for a neural network.  The first stacked channel has green color as in Figure~\ref{fig:channels} (a), and the second stacked channel has a magenta color as in Figure \ref{fig:channels} (b), and we already have the black background common in both images. To have the entire wing image, the green and magenta stacked channels were merged as in Figure \ref{fig:channels}. (c)
The dataset consists of stacked and combined microscopy images, each representing a snapshot of a Fruit fly wing at a specific time. The images are organized into 76 images of wings with tumors and 139 images without tumors. The tumor set includes images of different stages of cancer, including both early and late stages. However, some images in the no-tumor folder with a tiny tumor percentage will be suppressed in the coming days. 

\footnote{ The dataset: \url{https://doi.org/10.5281/zenodo.7648739}
    \label{Dataset}
} 

\section{Results and Discussion}
In this paper, we perform experiments to verify and validate Anna Karenina Principle (AKP). We create different families of models by using the  same Inceptionv3 architecture but with different initializers. All the Inceptionv3 model families are trained in similar environments, with perturbations through swapping. The perturbations (stress) divide the model class into two groups (i) happy families (models that have high accuracy) (i) unhappy families (models that have low accuracy). Models with high test accuracy have similar representations, and models with low test accuracy are dissimilar.  We also compare our results with Inceptionv3 model families and with the same initializer. We use stochastic gradient descent (SGD) and RMSprop as optimizers.

\subsection{Fixed Seed Experiments}

\subsubsection{Hot Swapping of activation functions}
The activation functions are changed at epochs 3, 6, 9, 12, 15, 18, and 21. RMSprop optimizer training accuracy shows the signs of  overfitting from the 5th epoch as shown in Figure~\ref{fig:FixedSeed} (a) and (e).  It is inferred that the hot swapping of activation functions resets the training process. The model loses parts of its training and tries hard to adapt its lost learning, causing overfitting. The similarity of the training curve across all the experiments validates our AKP principle: Models going through similar training dynamics behave similarly.  SGD optimizer training accuracy exhibited  more fluctuations than its RMSprop counterparts,  ranging from 45\% to 65\% as shown in \ref{fig:FixedSeed} (c). The model accuracy increases from 40\% accuracy  reached up to  60\%. After changing the activation function at epoch three from Tanh to Softplus, the model's accuracy dropped to 45\% and then increased again. The stochastic fluctuations of the accuracy continue until epoch 21. We stopped the swapping at Epoch 21 by fixing with Relu until the end of the training so the model could cool down and stabilize.  The validation accuracy of RMSProp is shown in Figure~\ref{fig:FixedSeed} (b). The validation accuracy started at  65\% and remained relatively stable until epoch 10, after which the accuracy fluctuated more and reached 80\% at the end of epoch 30. The influence of activation function swaps on the validation accuracy is minimal for SGD, as shown in  Figure~\ref{fig:FixedSeed} (d). SGD here has fewer fluctuations than RMSprop. Some of the  activation functions hot swapping using SGD show similar behavior starting with a validation accuracy of 35\% and gradually stabilizing from epoch 25 and settling down between the range of 60\% and 65\%. Though we discover most of the model families behave similarly (happy families), we discover an interesting effect in Figure~\ref{fig:FixedSeed} (d), where there is a bifurcation, and one model stays successful, and all others are not so successful.

\subsubsection{Interchangeability of loss functions}
\begin{figure*}
\centering
\subfigure[]{\includegraphics[width=0.24\linewidth,height=3cm]{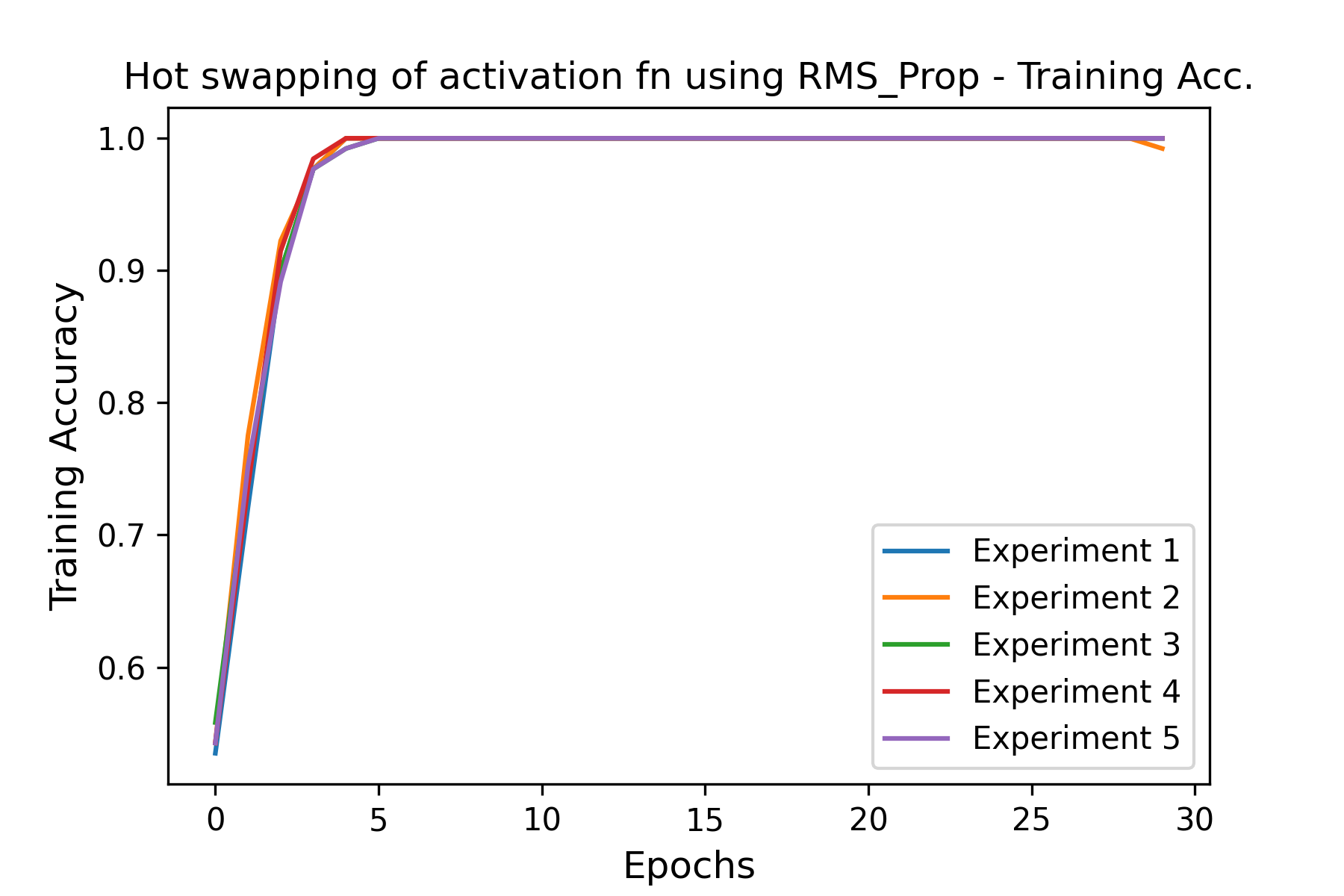}}
\subfigure[]
{\includegraphics[width=0.24\linewidth,height=3cm]{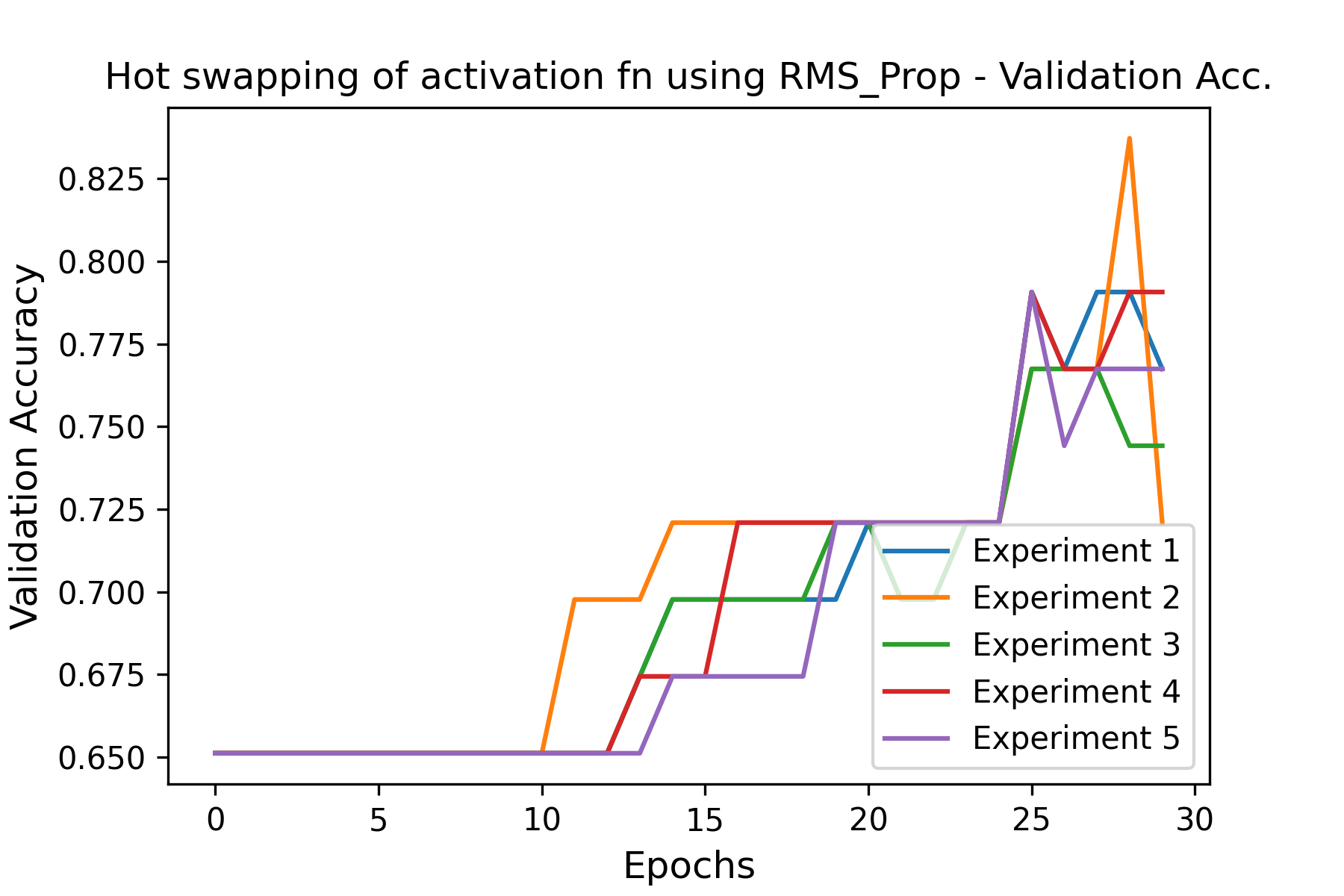}}
\subfigure[]{\includegraphics[width=0.24\linewidth,height=3cm]{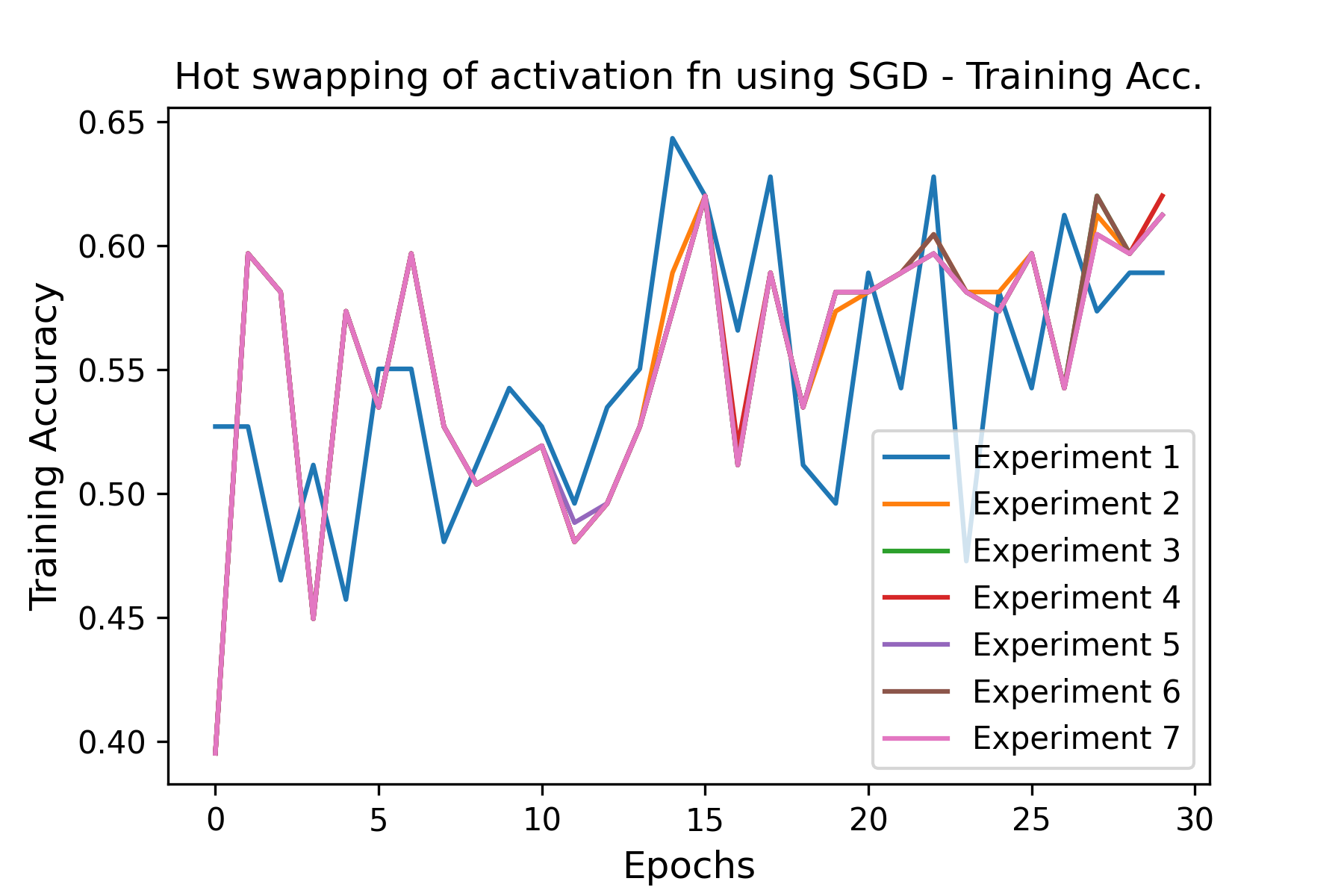}}
\subfigure[]{\includegraphics[width=0.24\linewidth,height=3cm]{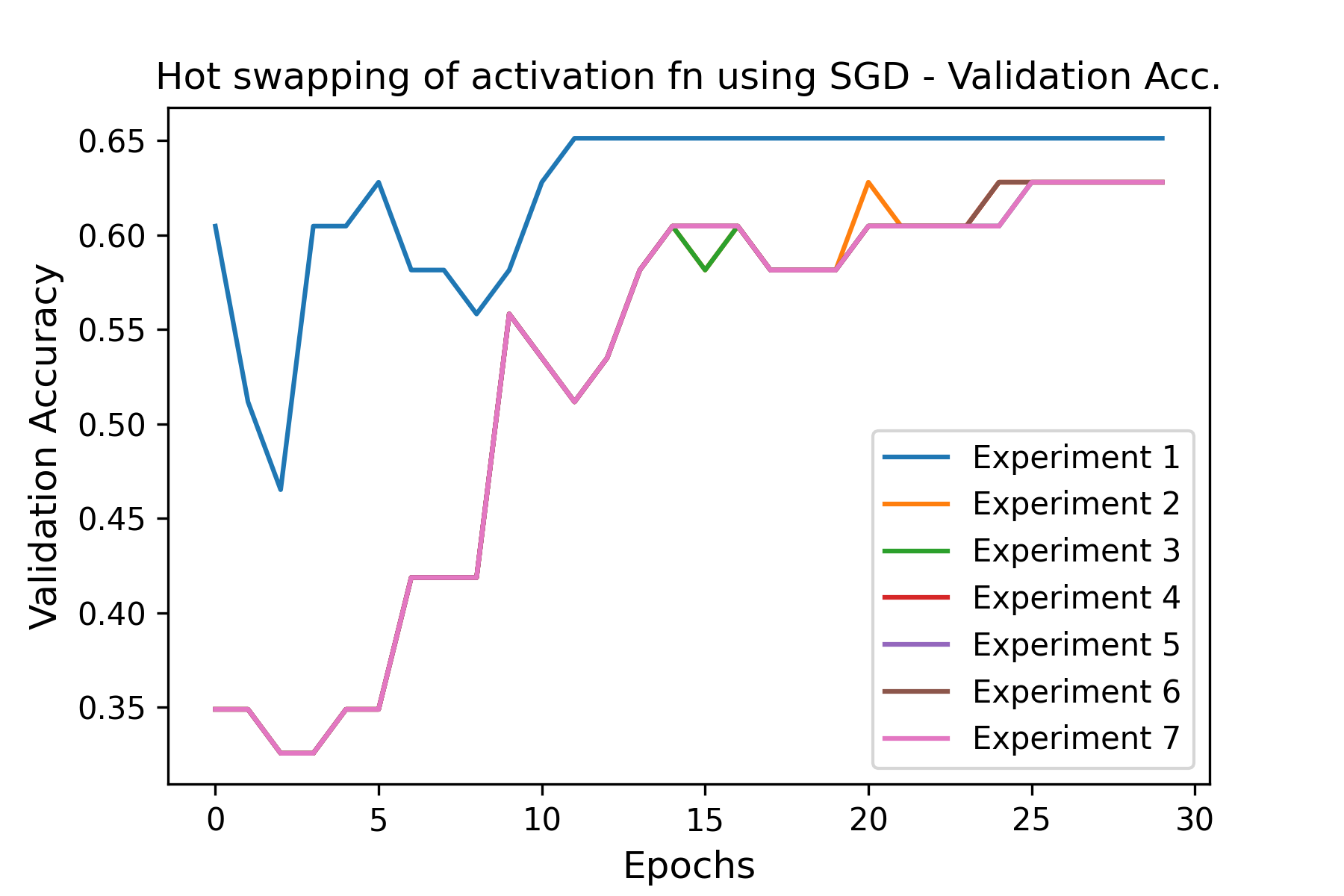}}
\subfigure[]{\includegraphics[width=0.24\linewidth,height=3cm]{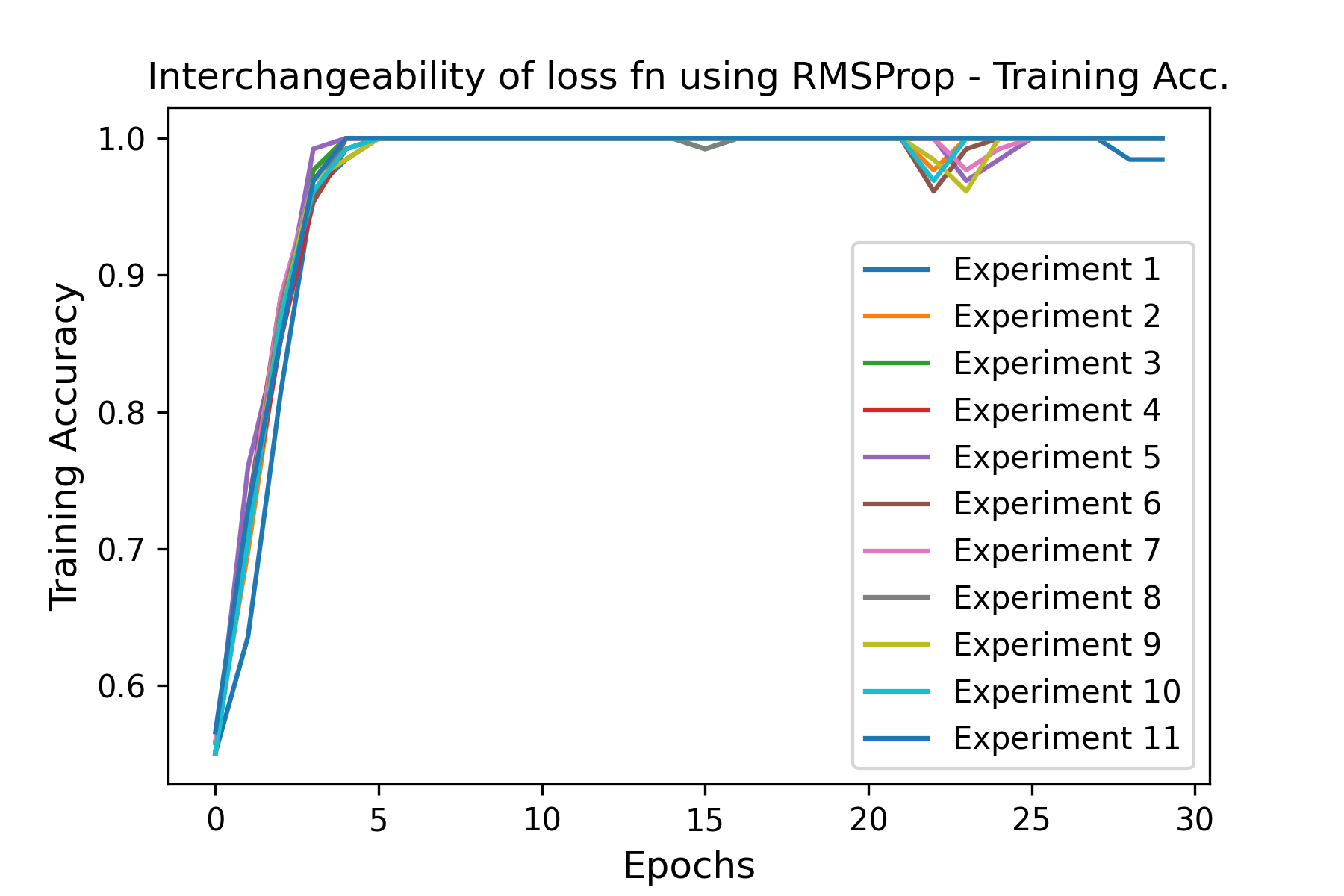}}
\subfigure[]{\includegraphics[width=0.24\linewidth,height=3cm]{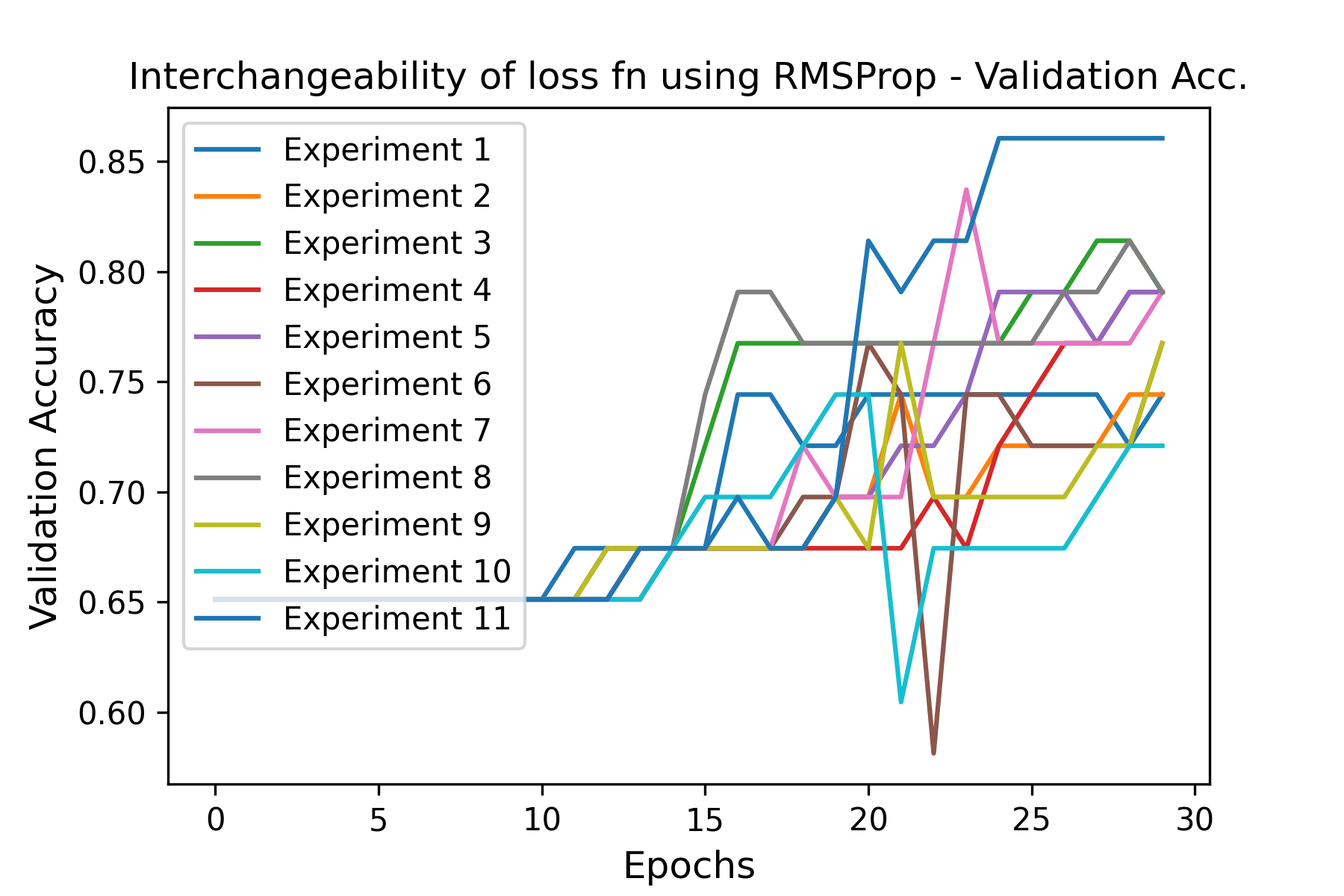}}
\subfigure[]{\includegraphics[width=0.24\linewidth,height=3cm]{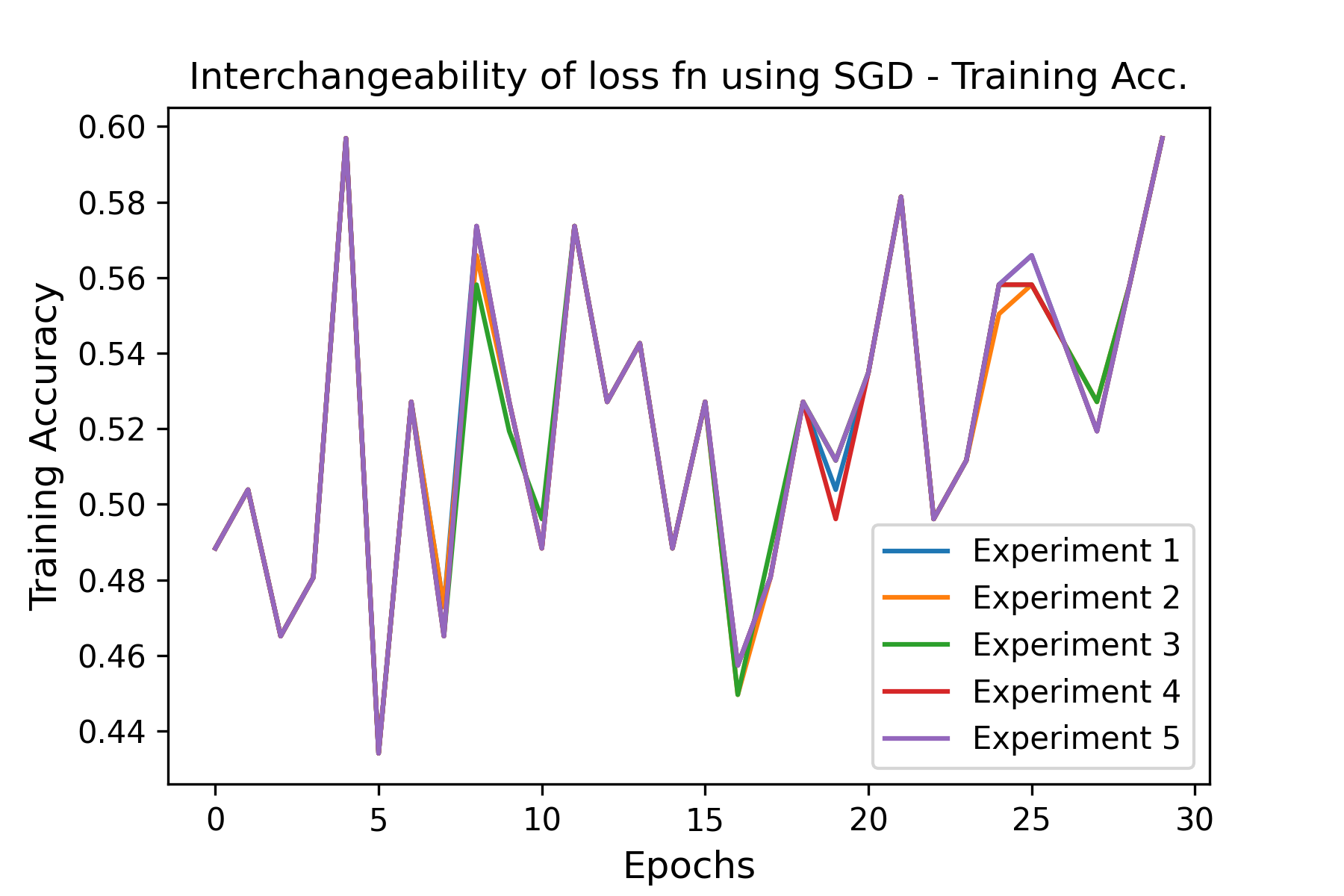}}
\subfigure[]{\includegraphics[width=0.24\linewidth,height=3cm]{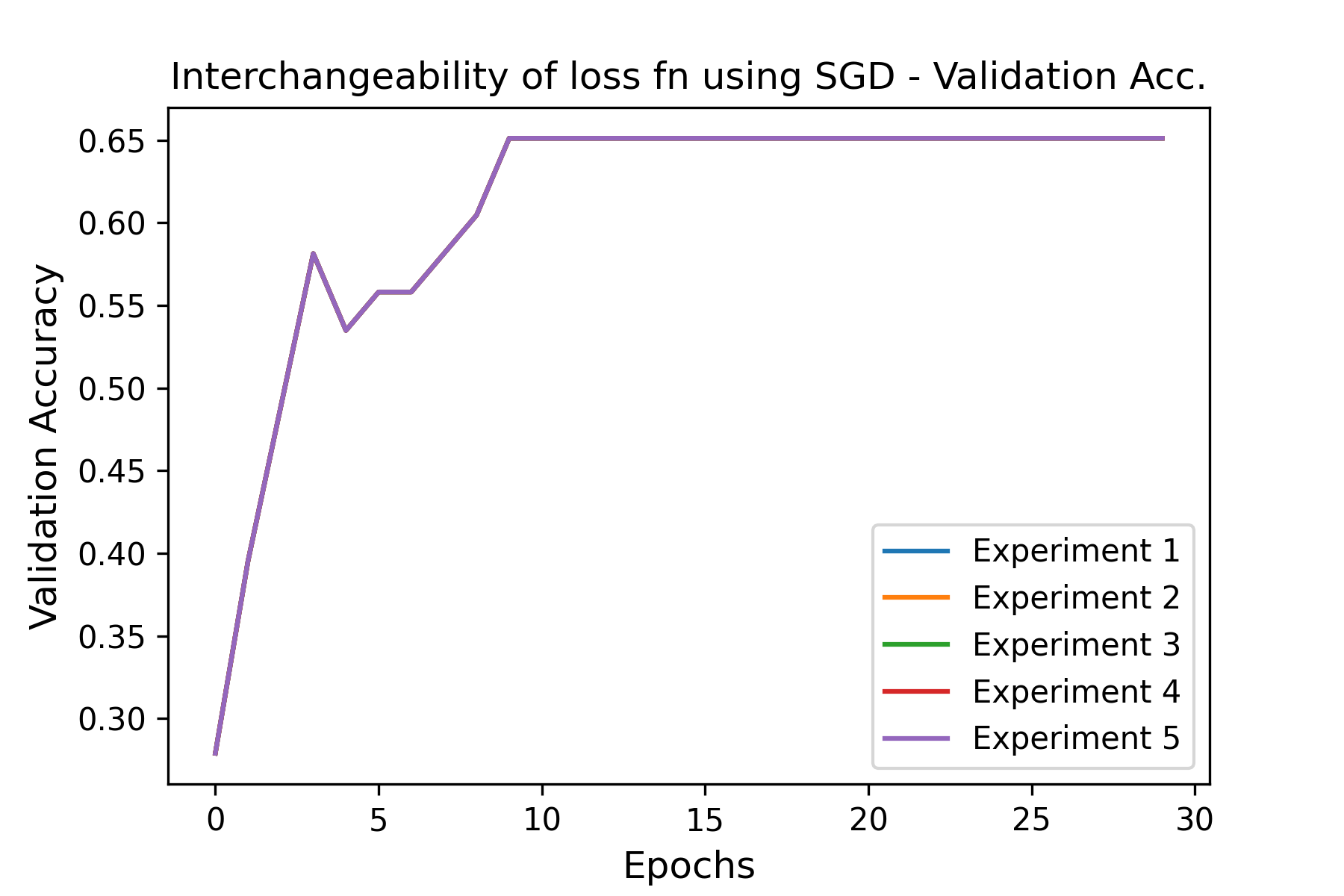}}
\caption{
Results of the Experiments using a Fixed Seed over 30 epochs (a) Experiments Training Accuracy of activation functions Hot swapping using RMSProp optimizer (b) Experiments Validation Accuracy of activation functions Hot swapping using RMSProp optimizer (c) Experiments Training Accuracy of activation functions Hot swapping using  SGD optimizer (d) Experiments Validation Accuracy of activation functions Hot swapping using  SGD optimizer (e) Experiments Training Accuracy of loss functions Interchangeability using RMSProp optimizer (f) Experiments Validation Accuracy of loss functions Interchangeability using RMSProp optimizer (g) Experiments  Training Accuracy of loss functions Interchangeability using SGD optimizer (h) Experiments Validation Accuracy of loss functions Interchangeability using SGD optimizer }
\label{fig:FixedSeed}
\end{figure*}

The training accuracy of RMSProp and SGD persisted in their behavior. RMSprop training accuracy stayed overfitting as shown in Figure \ref{fig:FixedSeed} (e). The training accuracy fluctuates more with SGD as shown in Figure~\ref{fig:FixedSeed} (g) between 44\% until 60\% . RMSProp Validation Accuracy in Figure~\ref{fig:FixedSeed} (f), however, started stable at 65\% until the 10th epoch, which exhibited fluctuations between 65\% and 80\%. In simpler terms, the model consistently struggled with overfitting during the training process, but despite this, it was still able to produce good results in terms of validation accuracy. SGD model achieved a stable validation accuracy of 65\% after eight training epochs, as shown in Figure \ref{fig:FixedSeed} (h). In summary, similarities and dissimilarities exist compared to the hot swapping of activation functions. The AKP is more pronounced with  SGD optimizer for both perturbations, though it is better with the  perturbations with interchangeability of loss functions.

\begin{figure*}
\centering
\subfigure[]{\includegraphics[width=0.24\linewidth,height=3cm]{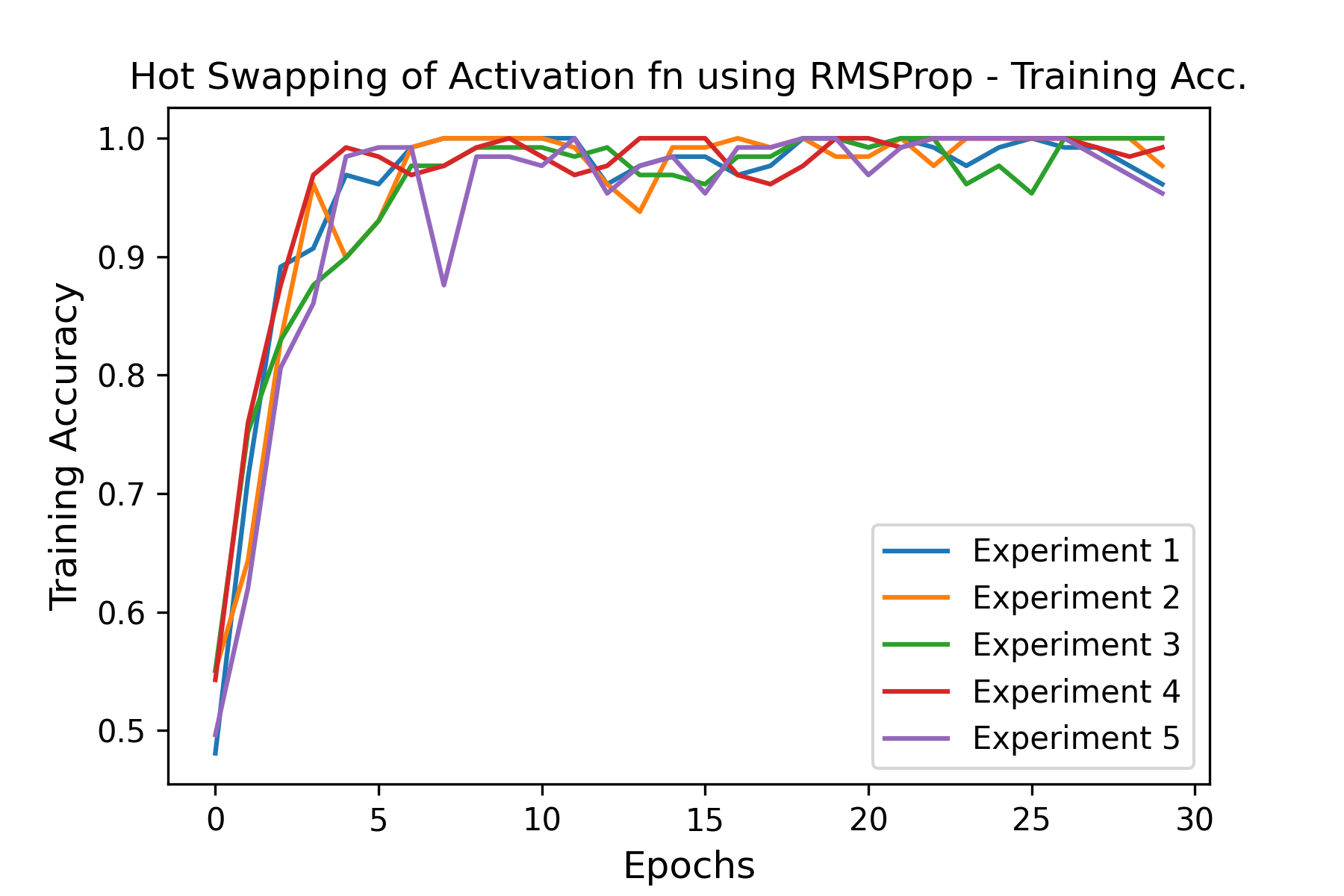}}
\subfigure[]
{\includegraphics[width=0.24\linewidth,height=3cm]{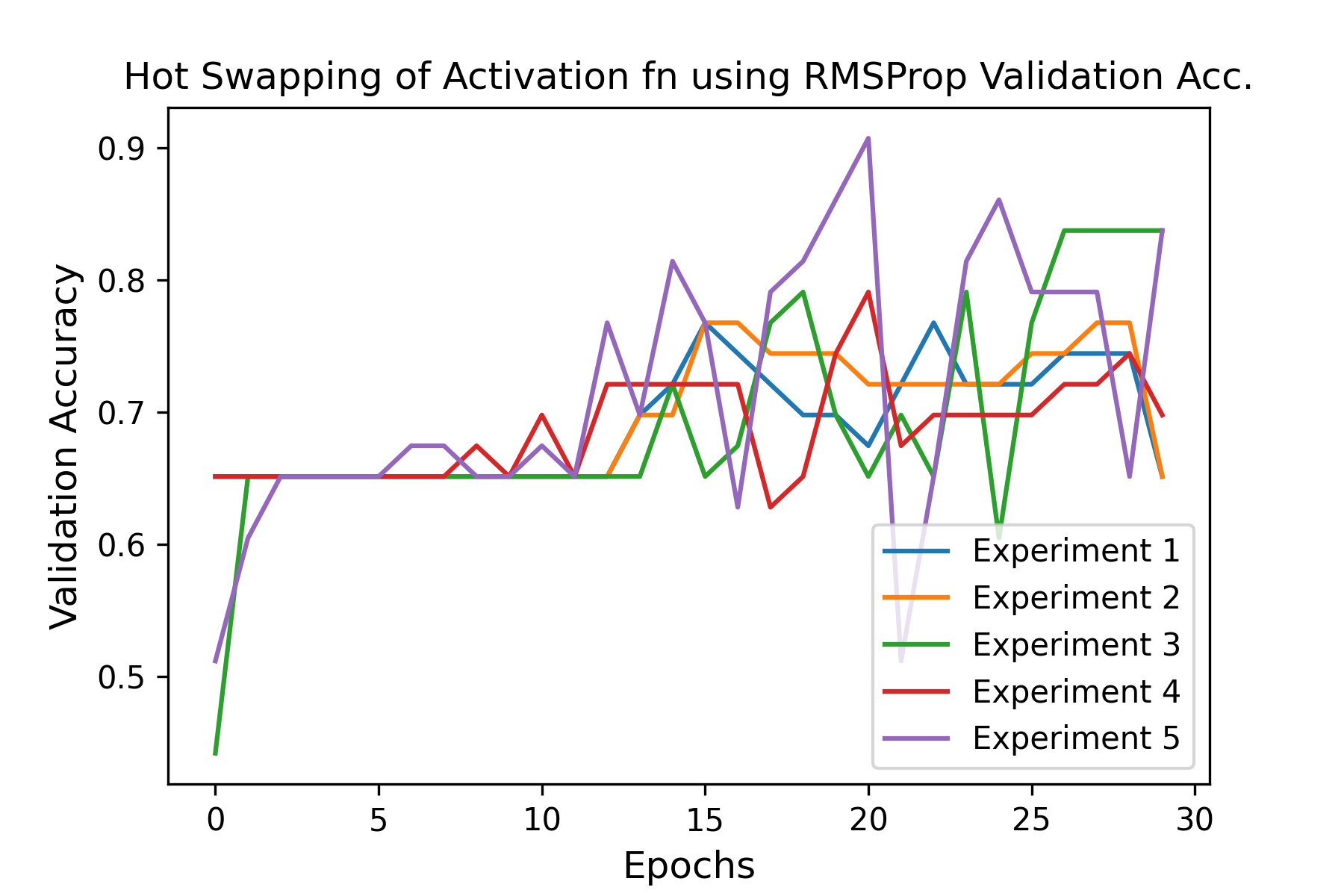}}
\subfigure[]{\includegraphics[width=0.24\linewidth,height=3cm]{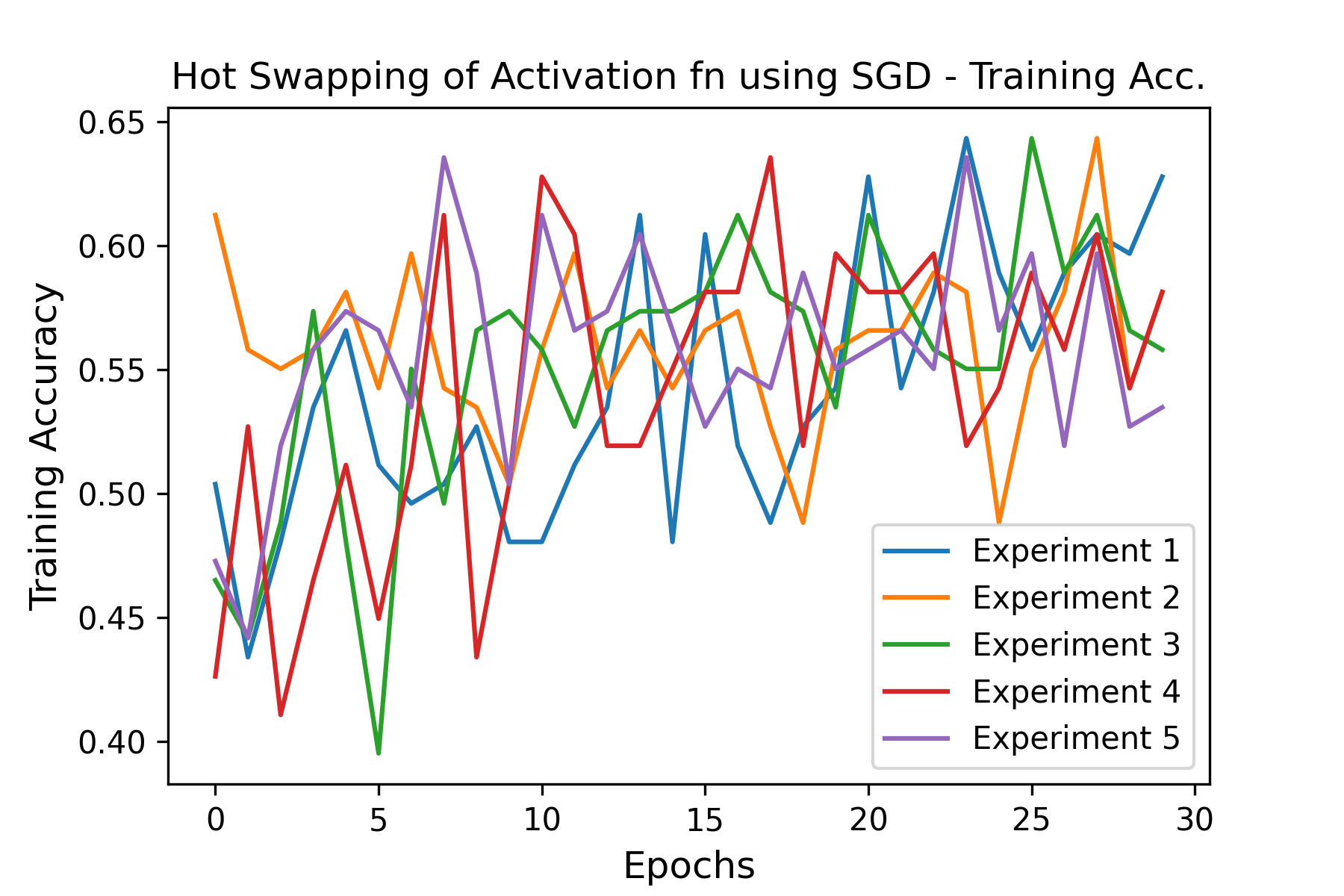}}
\subfigure[]{\includegraphics[width=0.24\linewidth,height=3cm]{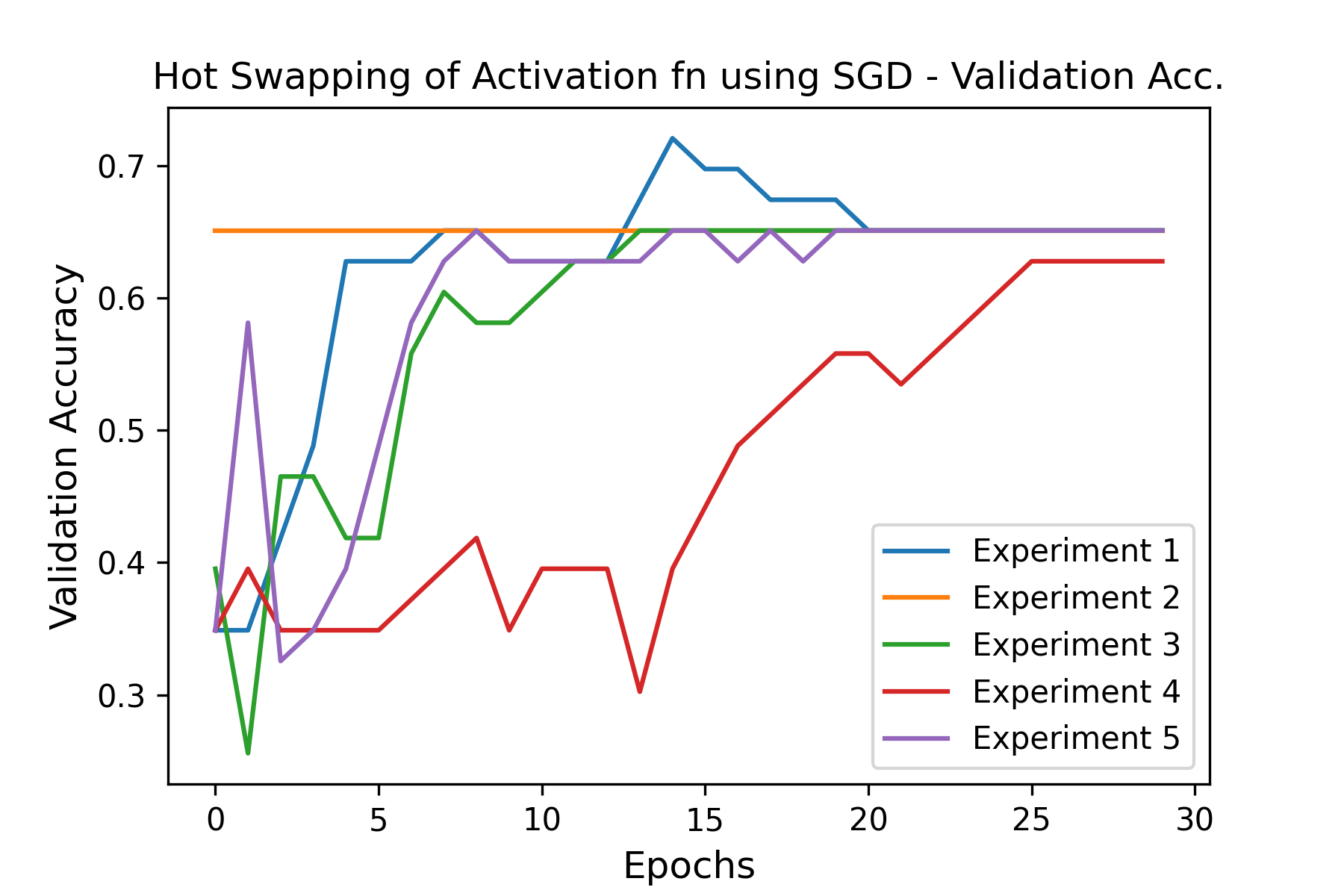}}
\subfigure[]{\includegraphics[width=0.24\linewidth,height=3cm]{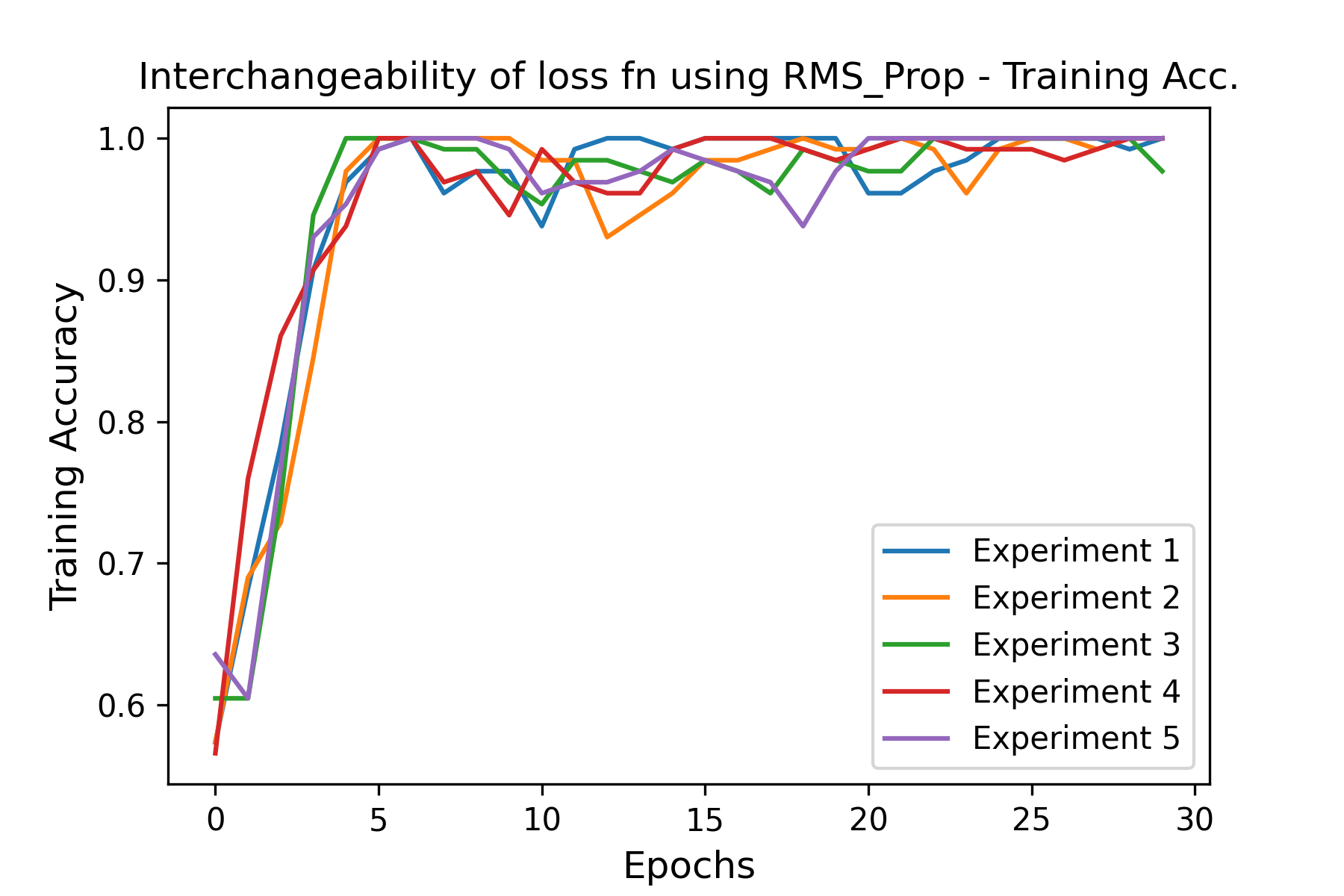}}
\subfigure[]{\includegraphics[width=0.24\linewidth,height=3cm]{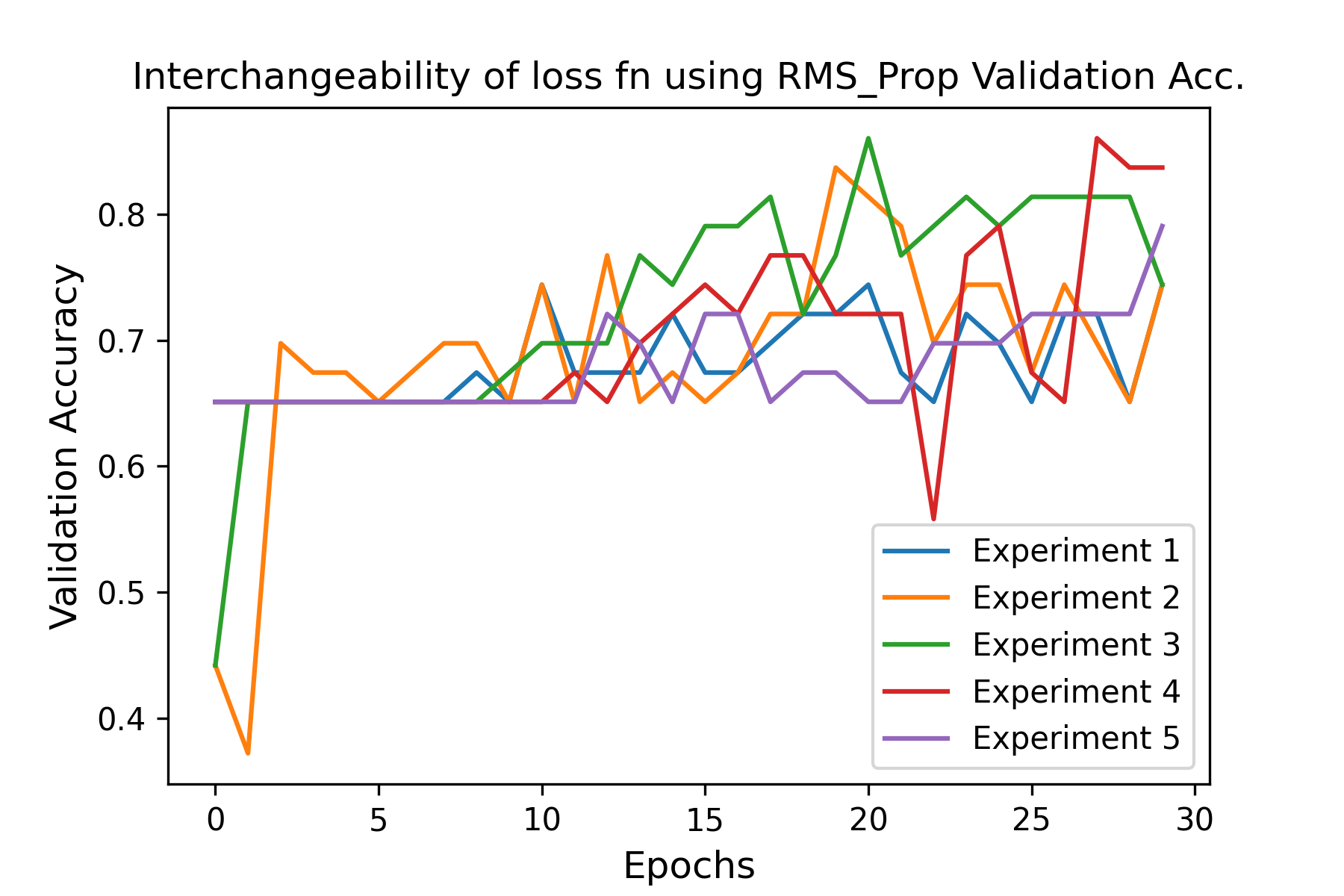}}
\subfigure[]{\includegraphics[width=0.24\linewidth,height=3cm]{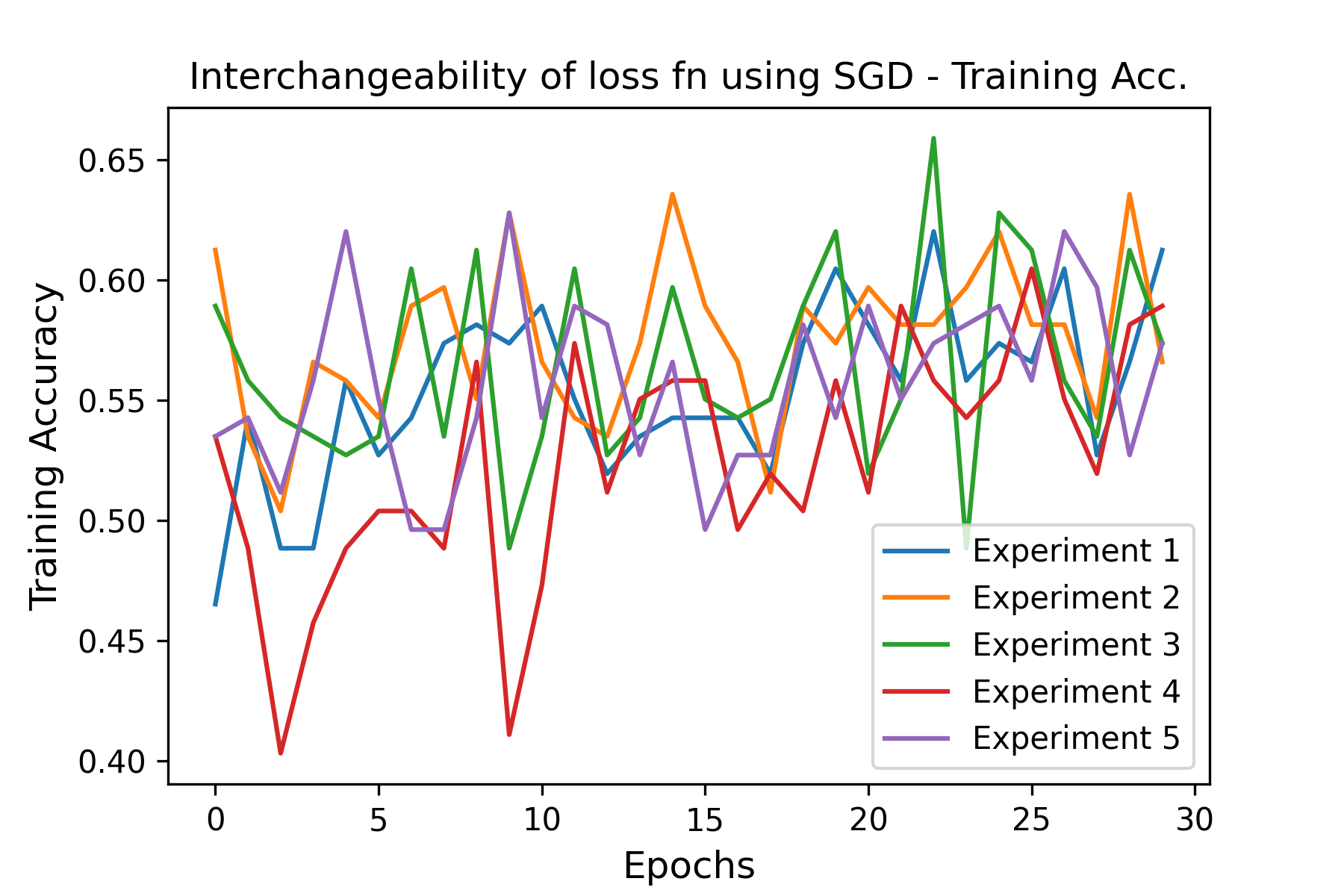}}
\subfigure[]{\includegraphics[width=0.24\linewidth,height=3cm]{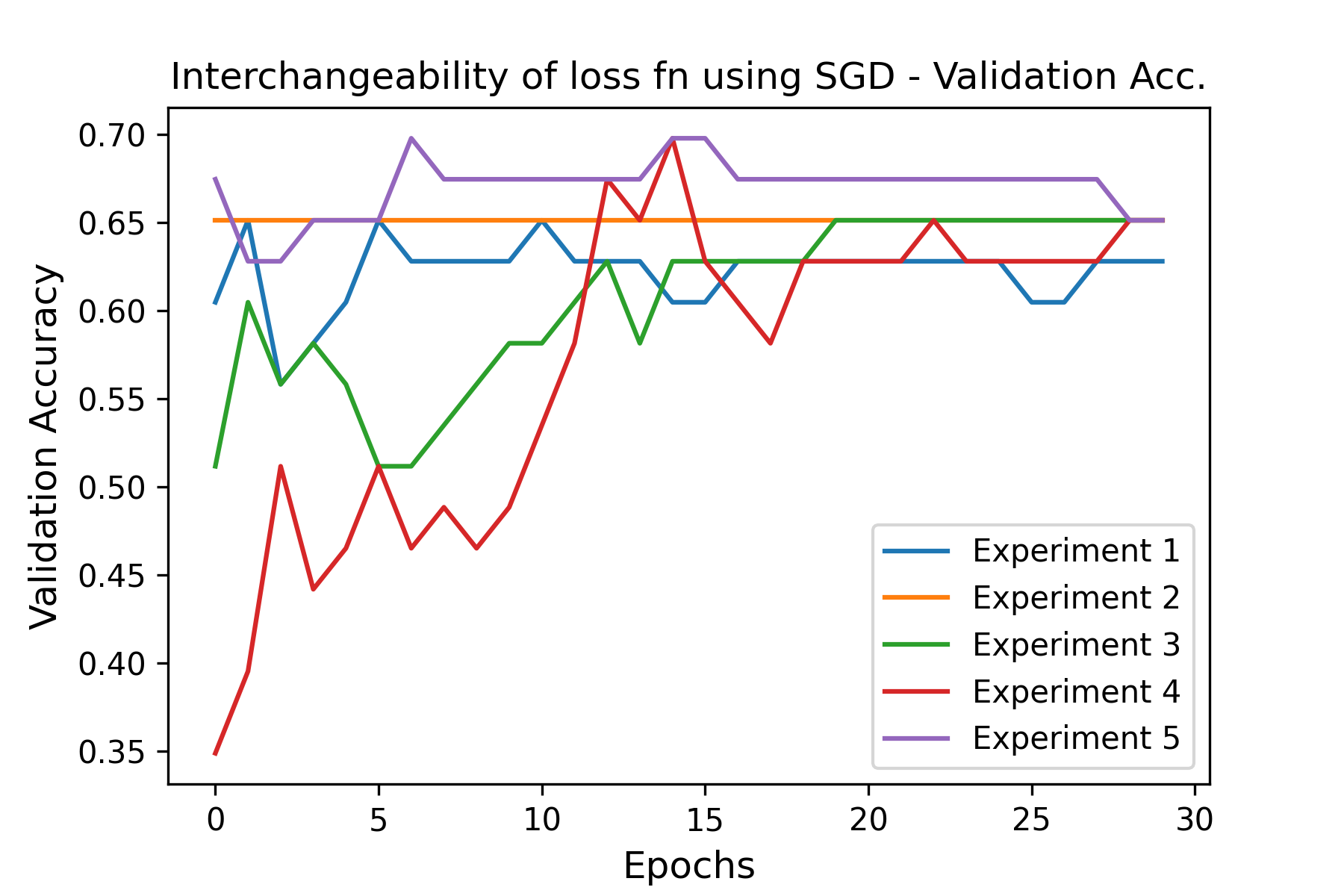}}
\caption{
Results of the Experiments using a Random Seed over 30 epochs  (a) Experiments Training Accuracy of activation functions Hot swapping using RMSProp optimizer (b) Experiments Validation Accuracy of activation functions Hot swapping using RMSProp optimizer (c) Experiments Training Accuracy of activation functions Hot swapping using  SGD optimizer (d) Experiments Validation Accuracy of activation functions Hot swapping using  SGD optimizer (e) Experiments Training Accuracy of loss functions Interchangeability using RMSProp optimizer (f) Experiments Validation Accuracy of loss functions Interchangeability using RMSProp optimizer (g) Experiments  Training Accuracy of loss functions Interchangeability using SGD optimizer (h) Experiments Validation Accuracy of loss functions Interchangeability using SGD optimizer }
\label{fig:RandomSeed}
\end{figure*} 

\begin{figure*}
\centering
\subfigure[]{\includegraphics[width=0.24\linewidth,height=3cm]{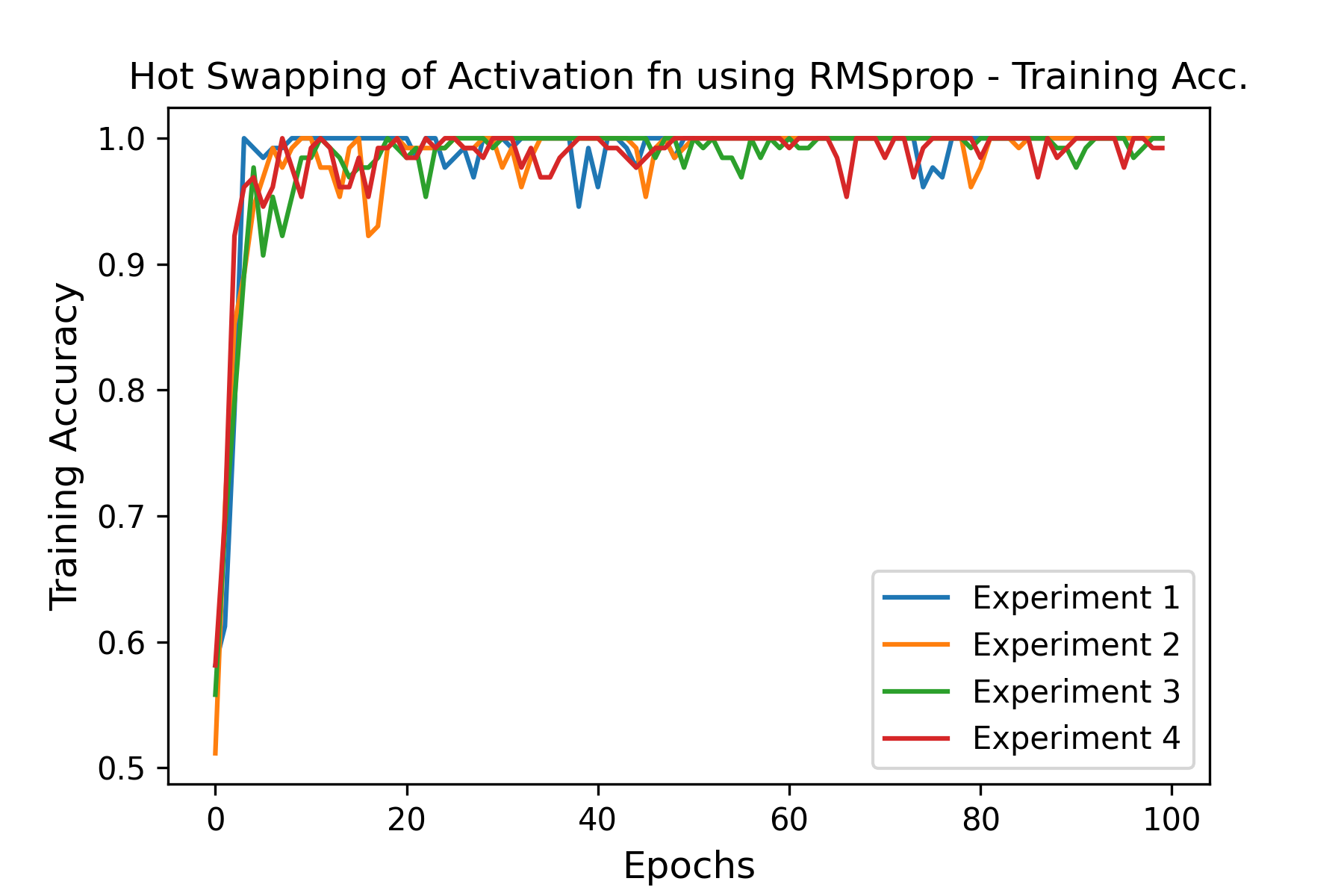}}
\subfigure[]
{\includegraphics[width=0.24\linewidth,height=3cm]{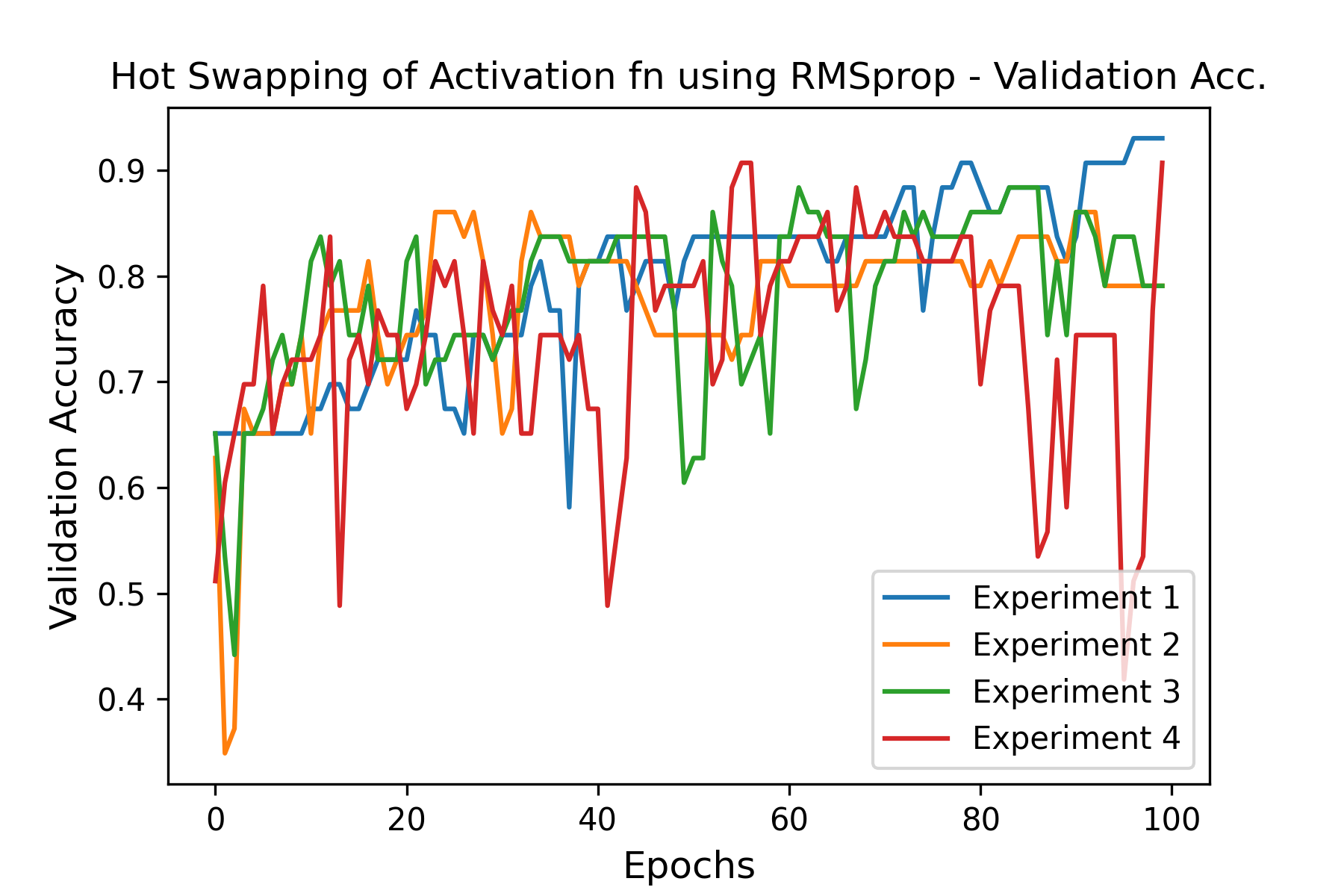}}
\subfigure[]{\includegraphics[width=0.24\linewidth,height=3cm]{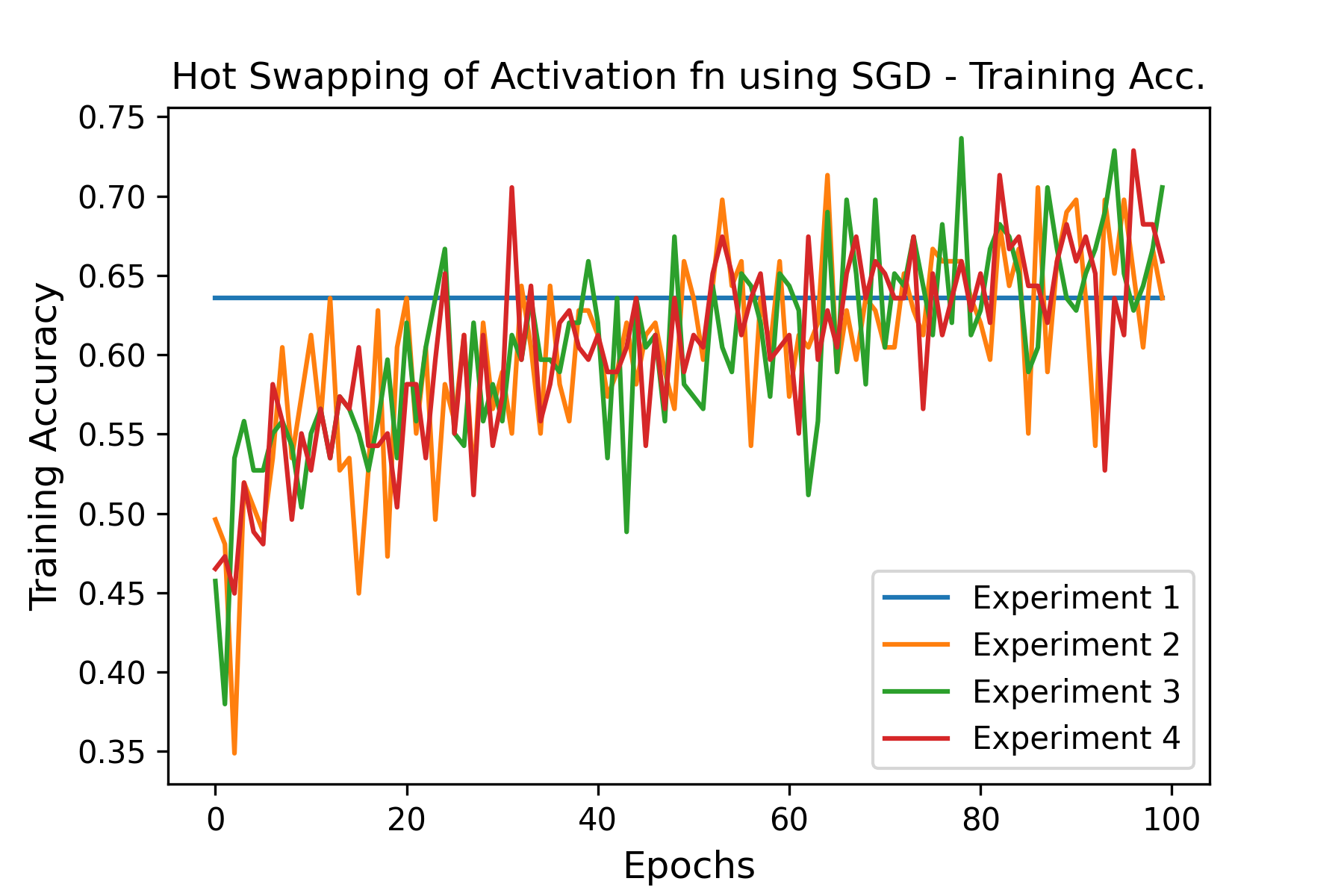}}
\subfigure[]{\includegraphics[width=0.24\linewidth,height=3cm]{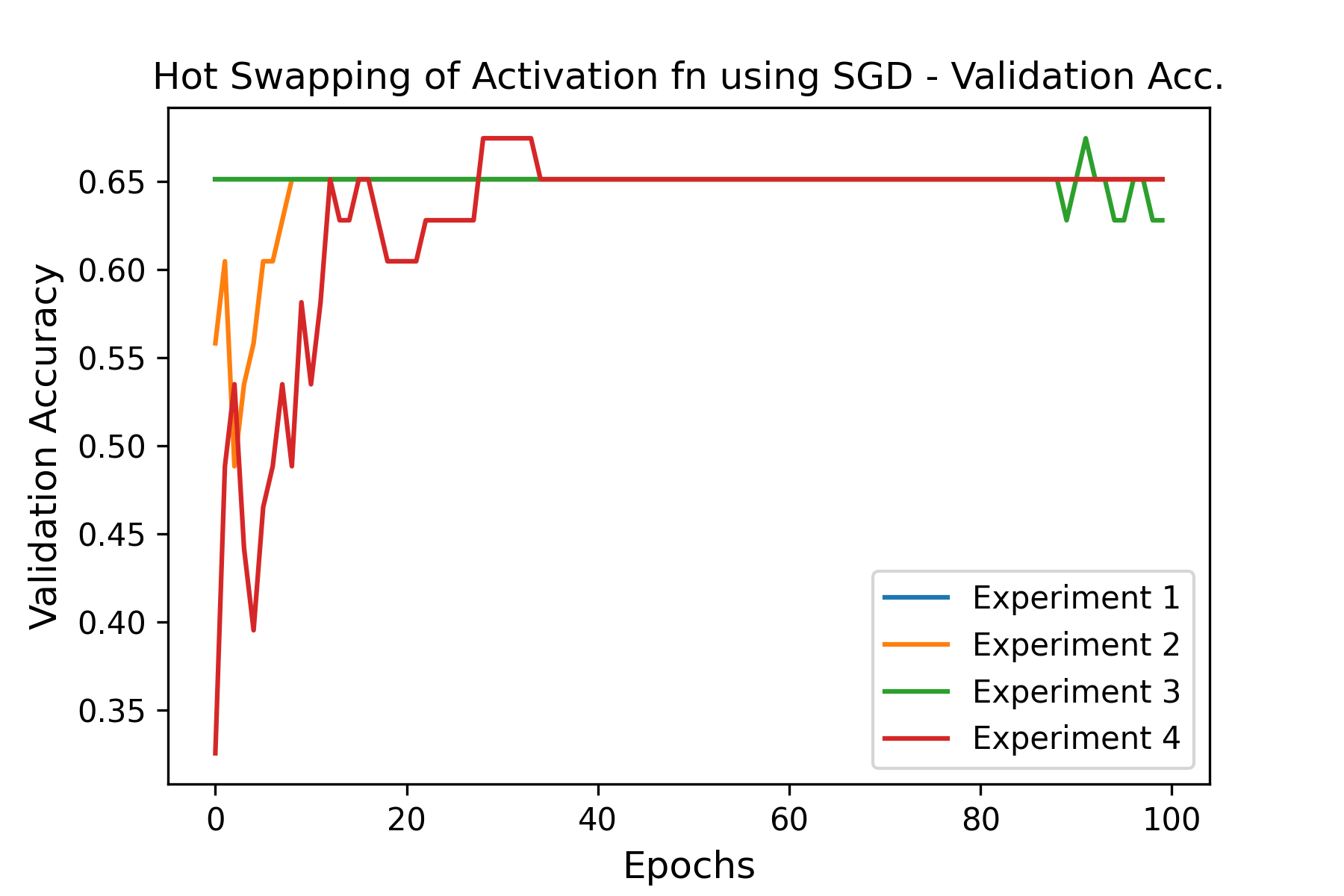}}
\subfigure[]{\includegraphics[width=0.24\linewidth,height=3cm]{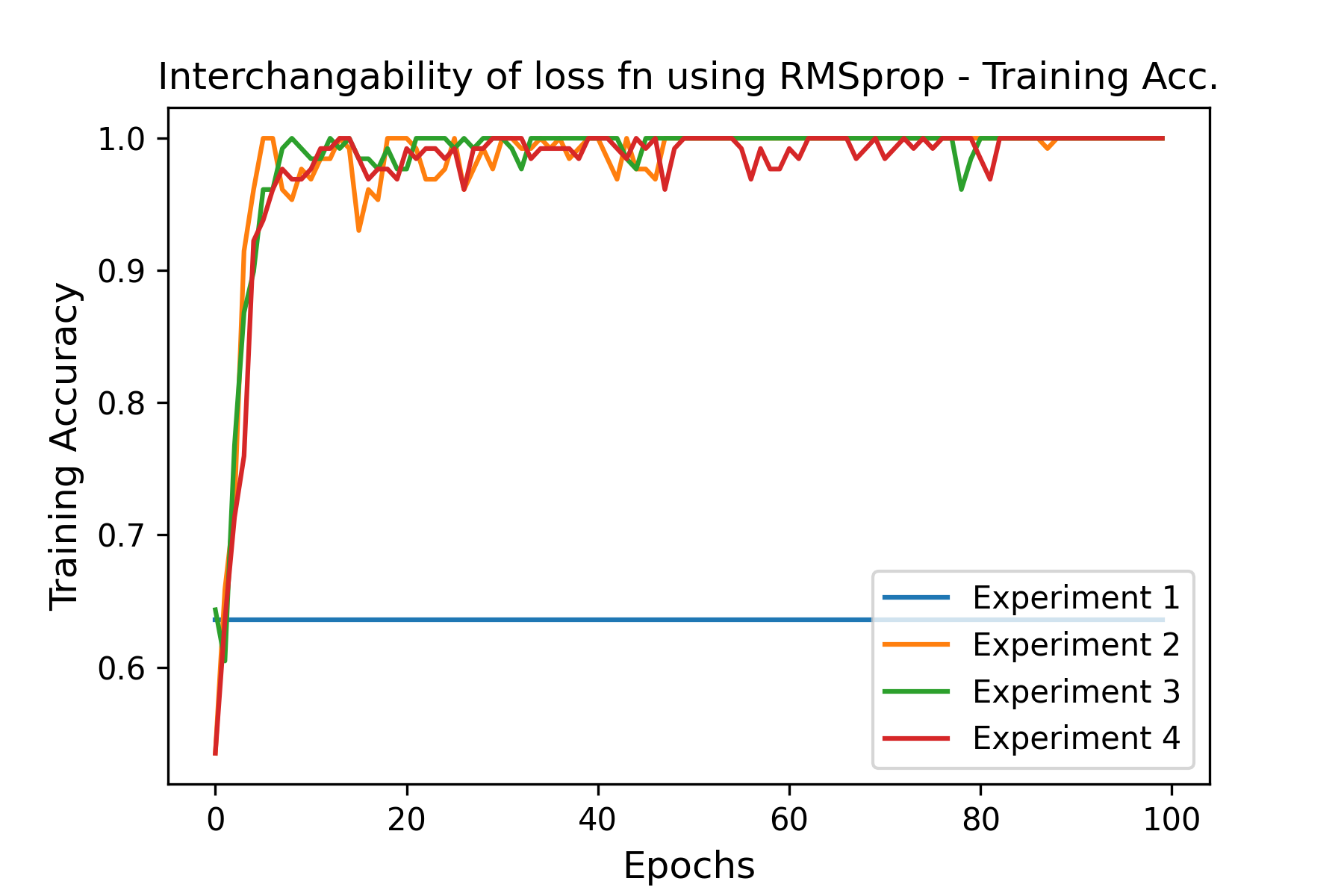}}
\subfigure[]{\includegraphics[width=0.24\linewidth,height=3cm]{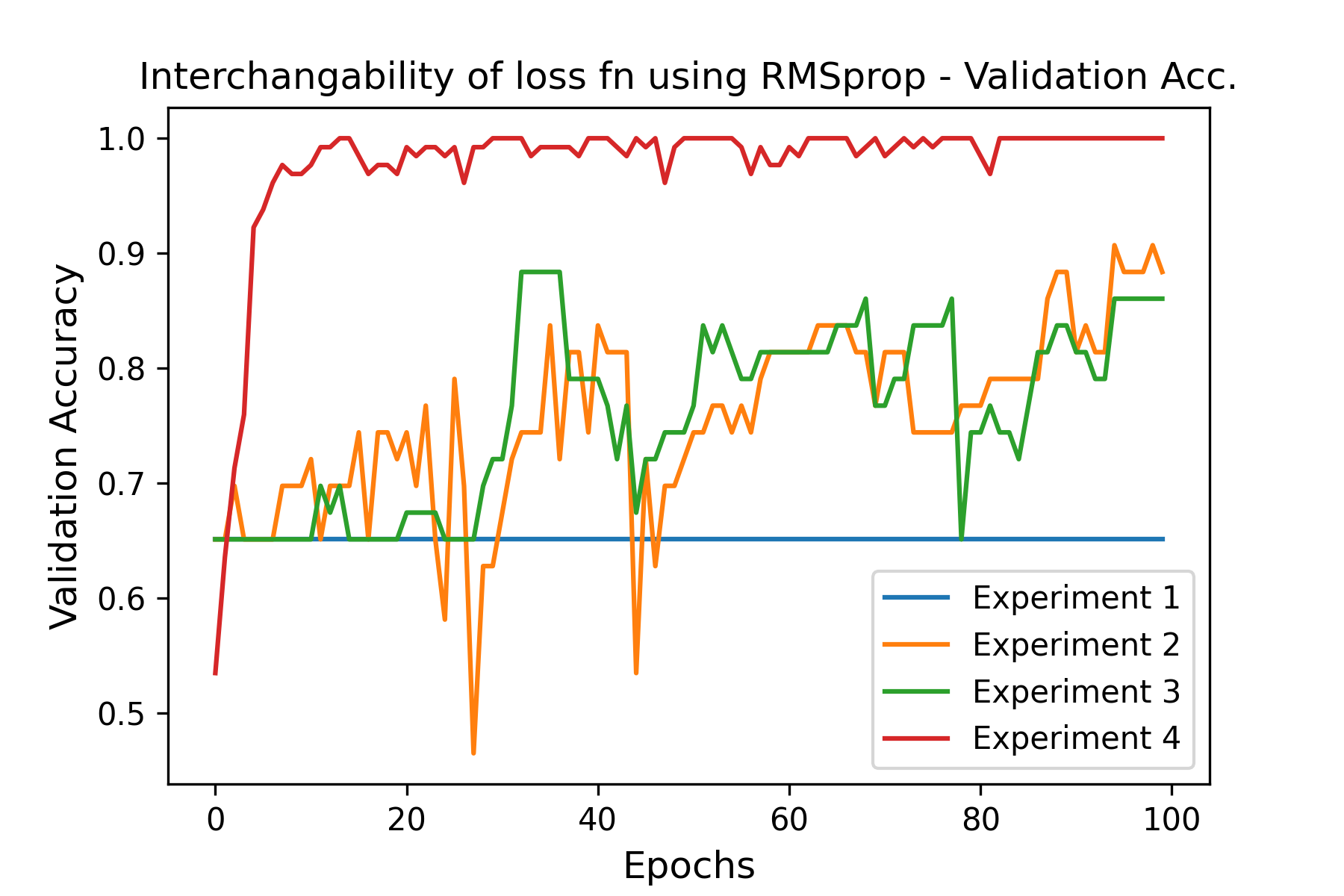}}
\subfigure[]{\includegraphics[width=0.24\linewidth,height=3cm]{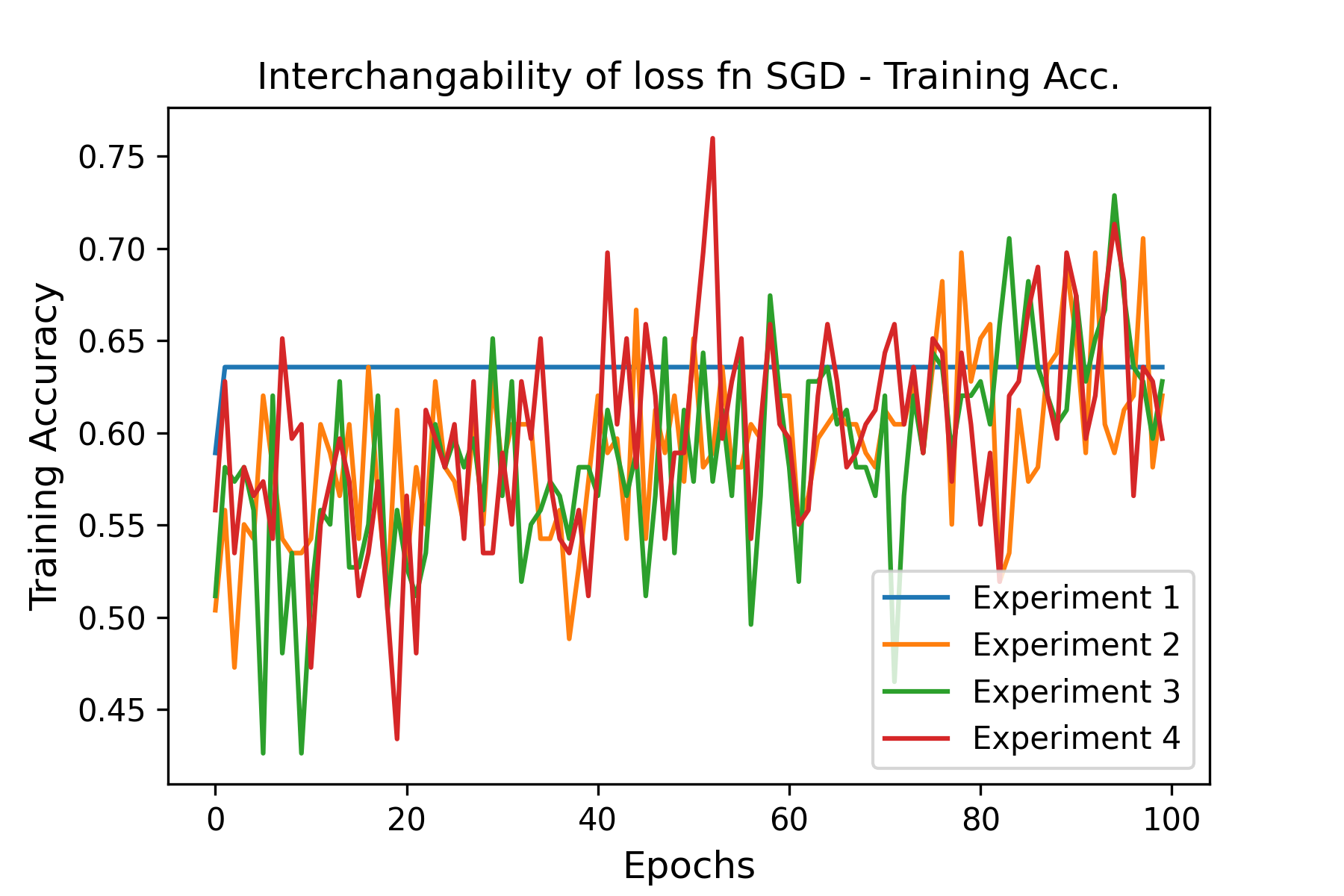}}
\subfigure[]{\includegraphics[width=0.24\linewidth,height=3cm]{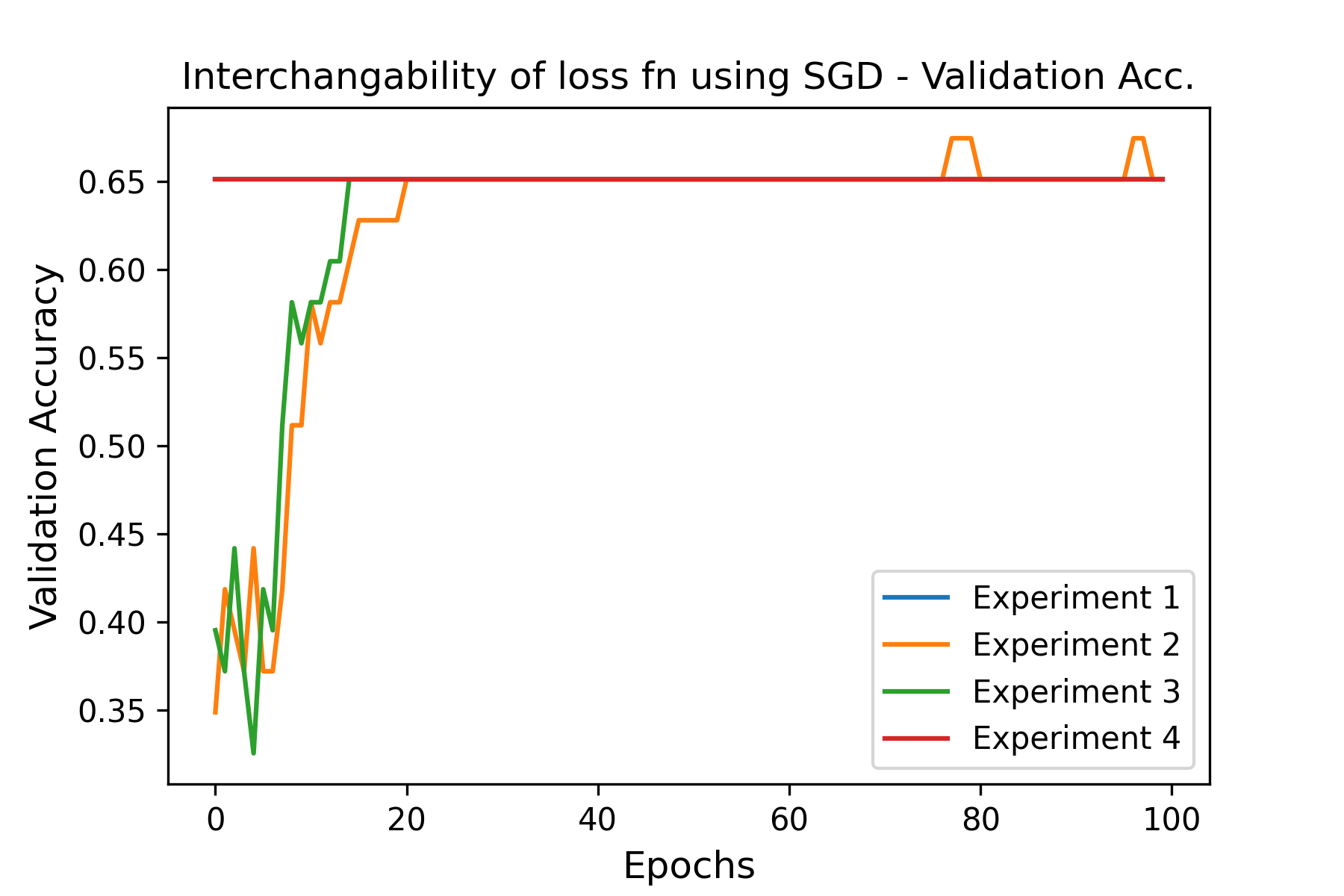}}
\caption{
Results of the Experiments using a Random Seed over 100 epochs  (a) Experiments Training Accuracy of activation functions Hot swapping using RMSProp optimizer (b) Experiments Validation Accuracy of activation functions Hot swapping using RMSProp optimizer (c) Experiments Training Accuracy of activation functions Hot swapping using  SGD optimizer (d) Experiments Validation Accuracy of activation functions Hot swapping using  SGD optimizer (e) Experiments Training Accuracy of loss functions Interchangeability using RMSProp optimizer (f) Experiments Validation Accuracy of loss functions Interchangeability using RMSProp optimizer (g) Experiments  Training Accuracy of loss functions Interchangeability using SGD optimizer (h) Experiments Validation Accuracy of loss functions Interchangeability using SGD optimizer }
\label{fig:Epoch100}
\end{figure*} 

\subsection{Random Seed Experiments}
\subsubsection{Hot Swapping of activation functions}
The training accuracy with the RMSProp optimizer has fewer fluctuations, but the overfitting trend is shown in Figure~\ref{fig:RandomSeed} (a). Evaluation accuracy for different families stayed together and started bifurcating from epoch 12 as shown in Figure~\ref{fig:RandomSeed} (b). The stress given to different models affects them differently. Some adapted well by picking the right features, and others could not. The training accuracy curve shows strong fluctuations with SGD optimizer ranging from 45\% to 65\% as shown in \ref{fig:RandomSeed} (c). Validation accuracy with SGD shows bifurcations where most models are happy, and the lone model is unhappy, as shown in Figure~\ref{fig:RandomSeed} (d). The stress affects the model differently. One model could not enlist itself to the happy member.

\subsubsection{Interchangeability of loss functions}
 A similar trend as in hot swapping of the activation functions is shown here in Figure~\ref{fig:RandomSeed} (e),(f),(g), and (h).

\subsection{Changing Kernel Initializers Experiments}
We conducted four experiments, each using a unique kernel initializer for the first dense layer of every model. The initializers employed were zeros, GlorotNormal, HeUniform, and TruncatedNormal initializer. For the TruncatedNormal initializer, we utilized parameters with a mean of zero and a standard deviation of 0.5.

\subsubsection{Hot Swapping of activation functions}
 The training accuracy of RMSProp has small fluctuations but persists in the overfitting trend as shown in Figure~\ref{fig:Epoch100} (a). In contrast, the SGD Training Accuracy for most experiments has fluctuations starting at 35\% and ending around 73\%. Except for experiment 1, using Zeros initializer  stayed at 63\%. The Validation Accuracy of RMSProp started at around 45\% and converged around 90\% as shown in Figure~\ref{fig:Epoch100} (b).
 
\subsubsection{Interchangeability of loss functions}
 Here the training accuracies of RMSProp also have small fluctuations and continue with the overfitting trend as shown in Figure~\ref{fig:Epoch100} (e), except Experiment 1 using Zeros initializer, the RMSProp Optimizer was unable to improve the training accuracy and stayed at 63\%. Experiment 1 using SGD shown in Figure ~\ref{fig:Epoch100} (g) also had the same training accuracy over the 100 epochs of 63\%. The validation Accuracy of SGD stayed at 65\% for the first experiment along all 100 epochs,  as shown in Figure~\ref{fig:Epoch100} (h)

\subsection{Test Classifications Accuracy}
Table \ref{table:1} presents the mean values of model predictions on the testing set. Each group of experiments underwent an average of five trials, except for Group C, where only four experiments were conducted. Analysis of the table shows that the SGD performs poorly in terms of precision, recall, and F-score for Class 1, which pertains to cancer images. 

\begin{table}[h!]
\tiny
\begin{center}
\begin{tabular}{|l|l|l|l|l|l|l|l|}
\hline
\# & Type & Optimizer & Class & Precision & Recall & F1 & Acc. \\
\hline
A & Hot Swapping Activation Fn & RMSProp & 0 & 0.8 & 1 & 0.88 &0.82 \\ 
& & & 1 & 1 & 0.47 & 0.64 &  \\ & & SGD & 0 & 0.57 & 0.50 & 0.52 & 0.43 \\ 
& & & 1 & 0.17 & 0.28 & 0.21  &\\ 
& Interchangeability of Loss Fn & RMSProp &
0 & 0.80 & 0.98 & 0.88 &0.82 \\
& & & 1 & 0.94 & 0.51 & 0.65 & \\ & & SGD & 0 & 0.67 & 1 & 0.81 & 0.67 \\
  & & & 1 & 0.0 & 0.0 & 0.0  &\\ 
\hline
B & Hot Swapping Activation Fn & RMSProp & 0 & 0.77 & 0.92 & 0.83 &0.75 \\
& & & 1 & 0.84 & 0.41 & 0.47  & \\ & & SGD & 0 & 0.63 & 0.78 & 0.66 & 0.58 \\
& & & 1 & 0.11 & 0.18 & 0.12  &\\ 
& Interchangeability of Loss Fn & RMSProp & 0 & 0.82 & 0.92 & 0.86 &0.80 
\\ & & & 1 & 0.87 & 0.57 & 0.64  
& \\ &  & SGD & 0 & 0.66 & 0.98 & 0.79 & 0.65 \\
  & & & 1 & 0.0 & 0.0 & 0.0  &\\ 
\hline
C & Hot Swapping Activation Fn & RMSProp & 0 & 0.82 & 1 & 0.90 &0.85 \\
  & & & 1 & 1 & 0.55 & 0.71 & \\  &  & SGD & 0 & 0.67 & 1 & 0.81 & 0.67 \\
  & & & 1 & 0.00 & 0.00 & 0.00  &\\ 
 & Interchangeability of Loss Fn & RMSProp & 0 & 0.80 & 0.96 & 0.88 &0.81 \\
  & & & 1 & 0.66 & 0.5 & 0.56  
& \\  &  & SGD & 0 & 0.67 & 1 & 0.81 & 0.67 \\
  & & & 1 & 0.25 & 0.01 & 0.03  &\\ 
\hline
\end{tabular}

\caption{Average Testing Accuracy of each group of experiments. Group A represents the Experiments with a Fixed seed over 30 epochs, Group B represents the Experiments with a Random Seed over 30 epochs, and Group C represents the Experiments with  Different Kernel Initializers over 100 epochs. The Table is reporting the Precision, Recall, F-score, and Testing Accuracy of the experiments, divided into class 0 [No Cancer], Class 1 [Cancer] }
\label{table:1}
\end{center}
\end{table}

\subsection{Discussion}
Although the RMSProp optimizer exhibits a similar pattern of over-fitting in the training accuracy, the  optimizer produced better validation and test accuracy results compared to the SGD  on our dataset. This suggests that while the RMSProp optimizer still struggles with over-fitting, it is better at generalizing to new unseen data, resulting in improved validation and test accuracy and the good performance of RMSProp due to its exponential averaging. The gradients of the perturbation noise decay with time, and this helps the  model to adapt. Leo Tolstoy's famous dictum that "all happy families look alike, each unhappy family is unhappy in its own way" can be applied to the results obtained in our experiments on training deep learning models. The stress in the neuron populations induces predator-prey dynamics, through which the features have evolved. 

Our results suggest that the models using the RMSProp optimizer as its adaptation strategy tend to have more consistent performance caused by the online hot swapping of activation and loss functions. On the other hand, the models using the SGD optimizer as its adaptation tools show poor performance compared to RMSprop. The fluctuations in the gradients get bigger because of no averaging in the  stochastic gradient descent. These fluctuations indicate each model's capacity to adapt to stress induced in the training process. This translates to test accuracy. The models using the RMSProp optimizer outperformed those using the SGD optimizer. We infer that even the overfitted models have good generalizable capacity.  RMSProp optimization methods may not be immune to overfitting, but they are still better able to generalize to unseen data, resulting in higher accuracy. We believe the training method through back-propagation can be improved by intertwining with some modified versions of lotka-volterra equation. These results suggest that RMSProp might be a more suitable optimization algorithm for our specific dataset. However, it is important to note that the optimal optimization algorithm can vary depending on the task and the dataset used.

\section{conclusion}
In this paper, we propose novel phenomena of the Anna Karenina principle (AKP) as a causal mechanism for the generalizability of deep neural models, specifically large models. The generalizability of deep learning remains a mystery, and there is no causal mechanistic theory to explain its generalizable bounds. Deep nets are even found to fit random labels~\cite{zhang2016understanding}. The probabilistic foundations can only partially explain the learning mechanism of deep models with billions of parameters. We propose an alternate learning theory of deep models from the principle of evolutionary games and population dynamics. The neurons are modeled as populations, and they are under evolutionary pressure to survive. The features of model learning come from the evolutionary pressure of neuron population. We translate nature's evolutionary pressure as perturbations of Inceptionv3 model families during its training. The pressure act as a lever, and different models evolve differently. Some towards good generalizable models and others towards poor generalizable models. All good generalizable models are happy families, and they share similar attributes, and all bad generalizable models (unhappy families) are unhappy in their own way. This is Anna Karenina Principle and is omnipresent from bacterial colonies to deep neural networks.

\bibliography{uai2023-template}
\section{Supplementary Material}
The high genetic similarity between humans and animals made it convenient to draw several research studies on animals and map the understanding to human genes to understand the development of diseases. About 75\% of the genes responsible for human diseases have similarities in flies \cite{ugur2016drosophila}.
Cancer deaths are increasing worldwide, and the community does not yet know how tumors grow and how to suppress them. There is a pressing need for experiments to understand the development of tumors in cells and how cells communicate with each other. There are extensive studies on the fruit fly, also known as drosophila, since genetics and tumor formation imaging are studied simultaneously. Reading the gene sequences is not the only way to detect Tumor suppressors and predict human cancer. Another way to understand the evolution of the tumor is to monitor the fly and notice how the cells evolve if it losses some proteins and its reaction to the increase of copies from specific proteins. Based on studies, Some proteins could help suppress cancer, while the loss of other proteins contributes a lot to the rise in cancer risk. Each gene in the cell contributes to the decision on what sort of cell it will be, its size, shape, and control tissue growth. One of the membrane proteins in drosophila is the scribble that maintains the apical-basal cell polarity of the epithelial tissues. A Recent Study showed that the loss of Scribble Protein or its partners from the tissue leads to abnormal growth of cells.
In contrast, the alpha-catenin protein expression in cells helps restore growth regulation and tumor suppressor activities. During the biological experiments over the drosophila wings, the researchers capture the evolution of the fly wing every day. In some of the experiments where the scribble was lost from some cells, it led to a tumor. In contrast, in others where the alpha-catenin protein was extensively expressed, the cell could suppress the tumor, and cells could return to normal without tumor cells.\cite{huang2022scribble}, With the advances in Computer vision and image processing using neural networks, it is possible to understand the evolution of cancer through the process. After training the neural network over the wing images, The model can predict whether a fly in the early stage will be a tumor or not.

The neural network can extract, learn and detect features from the images. It could find the most important features contributing the most to the decision of image classification. The neural network splits the image into small patches. The attention weights (priority) are given to these image patches. These attention weights represent the chance of cancer growth. The machine learns these attention weights (on its own) from the microscopy images by adjusting its parameter. This approach will allow us to build an accurate and transparent model, enabling healthcare providers to understand better and trust the predictions made by the system. Due to the considerable gap between the massive amount of images the neural network requires to train over and the small number of images generated in the experiments, Researchers adapt transfer learning \cite{weiss2016survey}. In this research,  transfer learning is used by customizing already pre-trained models to the new research target. 

\end{document}